\RequirePackage{fix-cm}
\documentclass[smallextended]{svjour3}       
\smartqed  
\usepackage{lineno,hyperref}
\usepackage{longtable}
\usepackage{graphicx}
\usepackage{subfigure}
\usepackage{multirow}
\usepackage{makecell}
\usepackage{colortbl}
\modulolinenumbers[5]
\usepackage{color,soul}

\usepackage{times}
\usepackage{epsfig}
\usepackage{graphicx}
\usepackage{amsmath}
\usepackage{amssymb}

\usepackage{multirow}
\usepackage{bm}
\usepackage{pifont}
\modulolinenumbers[5]
\begin{document}

\title{3D Point Cloud Descriptors in Hand-crafted and Deep Learning Age: State-of-the-Art
}


\author{Xian-Feng Han        \and
        Shi-Jie Sun \and
        Xiang-Yu Song \and
        Guo-Qiang Xiao
}


\institute{Xian-Feng Han \at
College of Computer and Information Science, Southwest University, Chongqing, 400715, China \\
              \email{xianfenghan@swu.edu.cn}           
           \and
            Shi-Jie Sun \at
            Chang'an University, Xi'an, ShaanXi,710064, China \\
            \email{hijiesun@chd.edu.cn}
            \and
            Xiang-Yu Song \at
            Deakin University,  VIC, 3125 Australia \\
            \email{xiangyu.song@deakin.edu.au}
            \and 
            Guo-Qiang Xiao \at
            College of Computer and Information Science, Southwest University, Chongqing, 400715, China
            \email{gqxiao@swu.edu.cn}
}

\date{Received: date / Accepted: date}

\maketitle

\begin{abstract}
The introduction of inexpensive 3D data acquisition devices has promisingly facilitated the wide availability and popularity of 3D point cloud, which attracts more attention to the effective extraction of novel 3D point cloud descriptors for accuracy of the efficiency of 3D computer vision tasks in recent years. However, how to develop discriminative and robust feature descriptors from  3D point cloud remains a challenging task due to their intrinsic characteristics. In this paper, we give a comprehensively insightful investigation of the existing 3D point cloud descriptors. These methods can principally be divided into two categories according to the advancement of descriptors: hand-crafted based and deep learning-based apporaches, which will be further discussed from the perspective of elaborate classification, their advantages, and limitations. Finally, we present the future research direction of the extraction of 3D point cloud descriptors.
\keywords{3D point cloud \and hand-craft descriptor \and deep learning based descriptor}
\end{abstract}

\section{Introduction}
\label{intro}
The availability of the low-cost 3D sensors, e.g., Microsoft Kinect, Prime Sense, has gained increasing interest in using three-dimensional point cloud \cite{Rusu20113D}\cite{Han2017A} for many applications in real scenarios, such as robot localization and navigation \cite{su2018splatnet}, autonomous driving, augmented reality \cite{wang2019associatively} and 3D medical imaging. Compared to  traditional 2D image, 3D point cloud has the capability of providing significantly richer information cues for analyzing objects and environments.

One considerably crucial step involved in these 3D applications aforementioned is the 3D descriptor or 3D feature extraction, which has a significant effect on the overall performance of descriptive result. Ideally, a robust and discriminative descriptor should be able to capture the geometric structure and be invariant to translation, scaling, and rotation at the same time. How to extract a meaningful 3D descriptor from  3D point cloud, however, is still a challenging research area that is worth being extensively investigated.

Specifically, the prosperously rapid development of deep learning technology in recent years has a remarkable impact on state-of-the-art achievements on various tasks related to object detection \cite{Girshick2015Fast}, semantic segmentation \cite{Shelhamer2014Fully} and visual recognition \cite{He2014Spatial}. Its great success leads to tend that increasing attention paying to 3D point cloud processing has shifted towards the deep learning-based period. As a result, the development of 3D point cloud descriptor is going through two major stages: hand-crafted period and deep learning-based period. The former kind of descriptors is mainly built on the basis of spatial and geometric attributes or relationships between points. While the latter uses deep neural networks to learn latent features from the 3D point cloud.  

\hl{A few papers have been published on 3D descriptors and its related fields, such as object recognition} \cite{Aldoma2012Tutorial}. \hl{However, (1) these papers} \cite{Alexandre20123D}\cite{Salti2012On}\cite{Mateo2014A}\cite{Guo2016A} \hl{covered only a rather limited number of descriptors to be evaluated. (2) Most works only concentrated on local-based descriptors for mesh or depth image while others focused on the performance of a small range of global descriptors for specific applications (e.g., Urban Object Recognition). (3) Particularly, among these studies, only one review }\cite{rostami2019a} \hl{available is related to deep learning. The authors reviewed the 3D shape descriptors from the machine learning point of view instead of their characteristics. And yet the coverage of 3D descriptors was not sufficient and this work was also lack of comprehensive performance summarization 
and comparison. Overall, there is no survey paper specifically emphasizing the comprehensive review, analysis, and evaluation of the 3D point cloud descriptors, especially these exploiting deep learning architectures at present.}

Compared with the previously mentioned research, the main contributions of this paper include:
\begin{enumerate}
\item \hl{To the best of our knowledge, this is the first review paper in the literature centering on advances in 3D point cloud descriptors, starting from traditional hand-crafted descriptors to deep learning-based approaches with significant research prospect. }

\item This is also the first survey paper taking into account the application of deep learning technology to the 3D point cloud. 

\item The main traits of these descriptors are summarized in table form, offering an intuitive understanding. 

\item A brief discussion of inspiring future research direction on 3D point cloud descriptors is present in this paper.

\end{enumerate}

The remainder of our paper is structured as follows: Section \ref{Tax} gives the taxonomy of 3D point cloud descriptors in this survey paper. Section \ref{hand} provides a comprehensive overview of the existing 3D point cloud descriptors in transitional hand-crafted age. The review of approaches in deep learning age is shown in Section \ref{sec:deep}. Section \ref{future} presents a discussion about the future research directions. Furthermore, the conclusion is drawn in Section \ref{con}.

\section{Taxonomy}
\label{Tax}
This paper mainly divides the 3D point cloud descriptors into two key categories in terms of feature extraction strategies, namely hand-crafted based and deep learning-based methods. 

For traditional hand-crafted approaches, the classification can be made by taking into account the region from which the descriptors are extracted (local, global and hybrid), the type of reference frame, and the attributes used (geometric information, spatial information, color information, other statistic information or combination of properties mentioned above.)

For deep learning-based descriptors, the representation of the input point cloud is crucial to the design of the neural network architecture. It also exerts the effect on computation cost and performance of point cloud-based applications. Particularly, the representation can be \textbf{(1)} intermediate feature, hand-crafted descriptor extracted from point cloud, \textbf{(2)} voxel grid, a kind of structured data, \textbf{(3)} multi-view, 2D images generated from point cloud, \textbf{(4)} Kd-tree, \textbf{(5)} Octree, two data structure to construct the topology, \textbf{(6)} point, point cloud itself as input, \textbf{(7)} graph, using graph structure to reflect the relationship between point and its neighbors and \textbf{(8)} multi-sensors. 

\begin{figure*}[!t]
\centering
\includegraphics[width=5in]{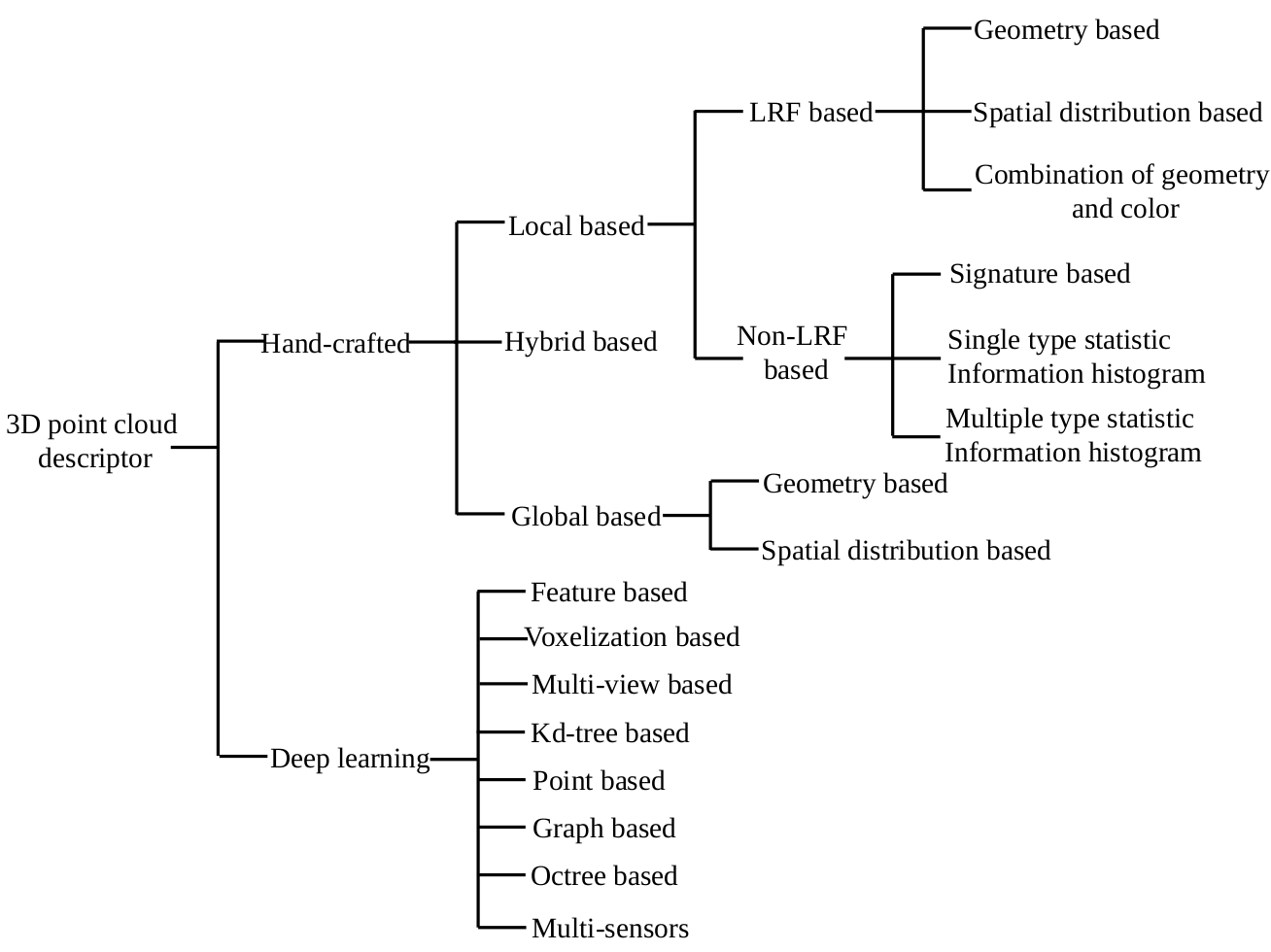}%
\caption{ The taxonomy of 3D point cloud descriptor.}
\label{fig_tax}
\end{figure*}

The detailed taxonomy of 3D point cloud descriptors is shown in Figure \ref{fig_tax}. And the detailed definition and introduction of these categories are shown in the following sections. 

\section{Dataset}
To evaluate the comprehensive performance of 3D point cloud descriptors for certain applications, datasets play greatly important roles in both traditional machine learning and deep learning era.
Table \ref{tab:dataset} presents a summary of the properties of datasets widely used by the state-of-the-art approaches, where some datasets are collected in the form of mesh instead of  point cloud. It is necessary, therefore, to introduce a preprocessing step to sample these models (such as ModelNet, ShapeNet) into discrete points.

\begin{table*}[h]
\caption{Some widely used datasets for 3D point cloud processing. "P": Part.}
\label{tab:dataset}
\resizebox{\textwidth}{!}{
\begin{tabular}{|c|c|c|c|c|c|c|c|c|c|}
\hline
NO. & Name                                                                            & Year & Type          & \#Categories & \#Objects & \#Models & \#Scenes  & Color & Scenario              \\
\hline
1   & Stanford 3D Models                                                              & 2003 & Synthetic     & 9            & 9         & 370      & -         & P     & Registration          \\
\hline
2   & Washington RGB-D Object                                                         & 2011 & Real          & 51           & 300       & 250,000  & -         & Y     & Ojbect Classification \\
\hline
3   & RGB-D Stereo Object Category \cite{Marianna2012From}           & 2012 & Real          & 14           & 14        & 140      & 10        & Y     & Object Classification \\
\hline
4   & Sydney Urban Objects Dataset \cite{de2013unsupervised}         & 2013 & Real          & 26           & 26        & 631      & -         & N     & Object Classification \\
\hline
5   & BigBIRD Dataset \cite{singh2014bigbird}                        & 2014 & Real          & -            & 125       & 75,000   & -         & Y     & Object Classification \\
\hline
6   & ShapeNet \cite{Chang2015ShapeNet}                              & 2015 & Synthetic     & 55           & 55        & 51,300   & -         & N     & Object Classification \\
\hline
7   & ModelNet \cite{wu20153d}                                       & 2015 & Synthetic     & 662          & 662       & 127,915  & -         & N     & Object Classification \\
\hline
8   & Oxford Radar Robotcar Dataset \cite{RadarRobotCarDatasetArXiv} & 2019 & Real, Outdoor & -            & -         & -        & 2,405,785 & Y     & Autonomous Driving    \\

\hline
9   & S3DIS \cite{2017arXiv170201105A}                               & 2017 & Real,Indoor   & 13           & 13        & -        & 6         & Y     & Segmentation          \\
\hline
10  & ScanNet \cite{dai2017scannet}                                  & 2017 & Real,Indoor   & 21           & 21        & -        & 1,513     & Y     & Segmentation          \\
\hline
11  & Semantic3D \cite{hackel2017isprs}                              & 2017 & Real, Outdoor & 8            & 8         & -        & 30        & Y     & Segmentation \\
\hline

12 & SemanticKITTI \cite{behley2019semantickitti:} & 2019 & Real, Outdoor& 28 & 28 & - & 23,201 & Y & Segmentation \\
\hline
\rowcolor{yellow}
13 & Oakland3d \cite{Munoz-2009-10227} & 2009 & Real, Outdoor & 44 & 44 & - & 17 & Y & Segmentation and Classification \\
\hline
\rowcolor{yellow}
14 & Paris-Lille-3D \cite{Xavier2018Paris} & 2018 & Real, Outdoor & 50 & 50 & - & 3 & Y & Segmentation and Classification \\
\hline

\end{tabular}}
\end{table*}

\section{3D point cloud Descriptors in hand-crafted age and discussion}
\label{hand}
Due to the specific characteristics of 3D point cloud data, hand-crafted descriptors play a major role in approaches directly processing the 3D point cloud. The existing 3D descriptors in literature can be distinctly classified into three primary categories:
global-based descriptors, local-based descriptors, and hybrid-based descriptors. 

\begin{figure*}[!t]
\centering
\subfigure[]{\includegraphics[width=0.5in]{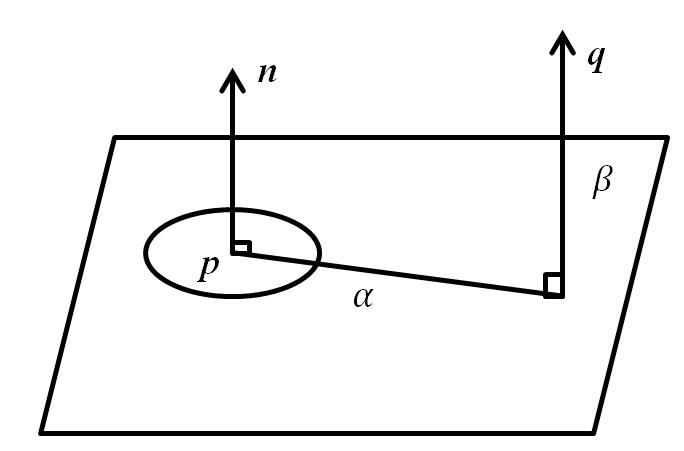}%
\label{fig_SI}}
\hfil
\subfigure[]{\includegraphics[width=0.5in]{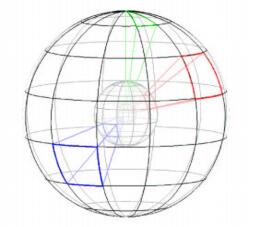}%
\label{fig_3DSC}}
\hfil
\subfigure[]{\includegraphics[width=0.5in]{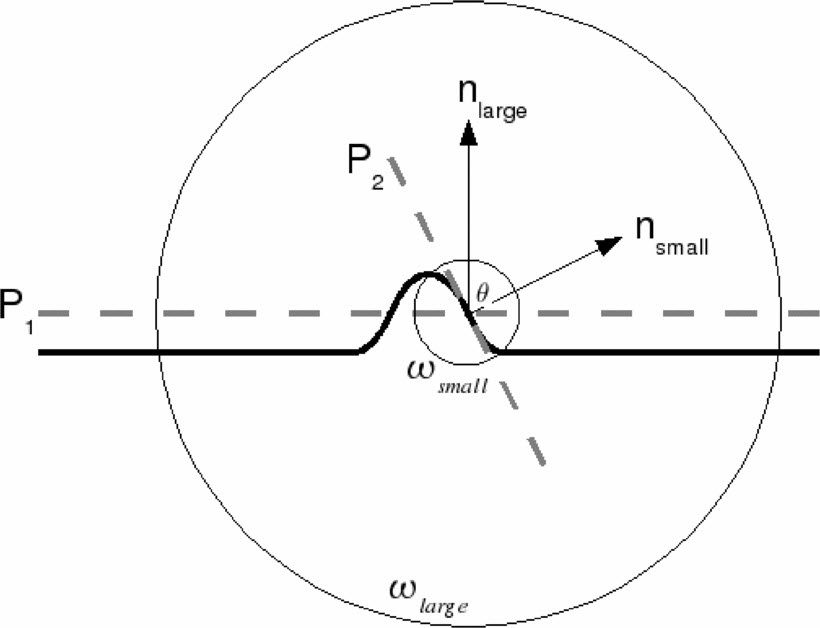}%
\label{fig_Thrift}}
\hfil
\subfigure[]{\includegraphics[width=0.5in]{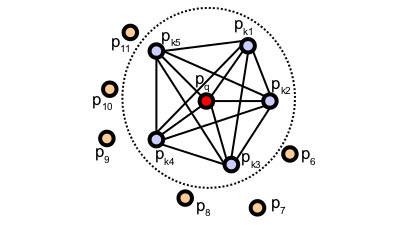}%
\label{fig_pfhdiagram}}
\hfil
\subfigure[]{\includegraphics[width=0.5in]{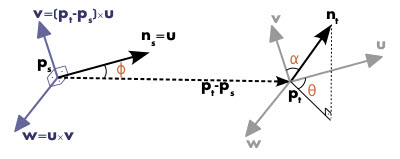}%
\label{fig_pfhframe}}
\hfil
\subfigure[]{\includegraphics[width=0.5in]{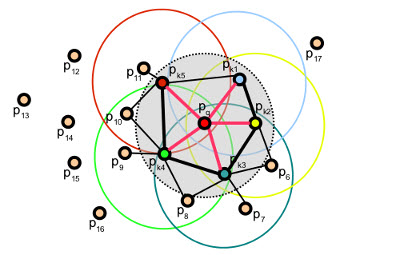}%
\label{fig_fpfhframe}}

\subfigure[]{\includegraphics[width=0.5in]{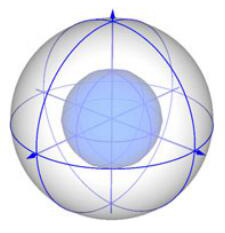}%
\label{fig_shot}}
\hfil
\subfigure[]{\includegraphics[width=0.5in]{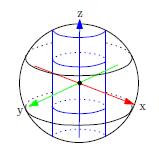}%
\label{fig_sh}}
\hfil
\subfigure[]{\includegraphics[width=0.5in]{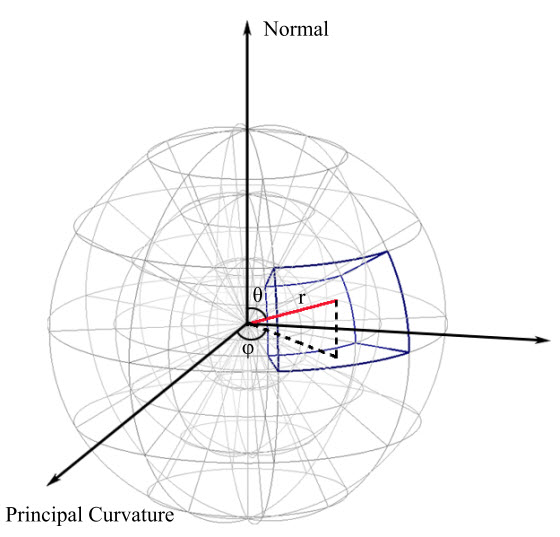}%
\label{fig_3DSSIM}}
\hfil
\subfigure[]{\includegraphics[width=0.5in]{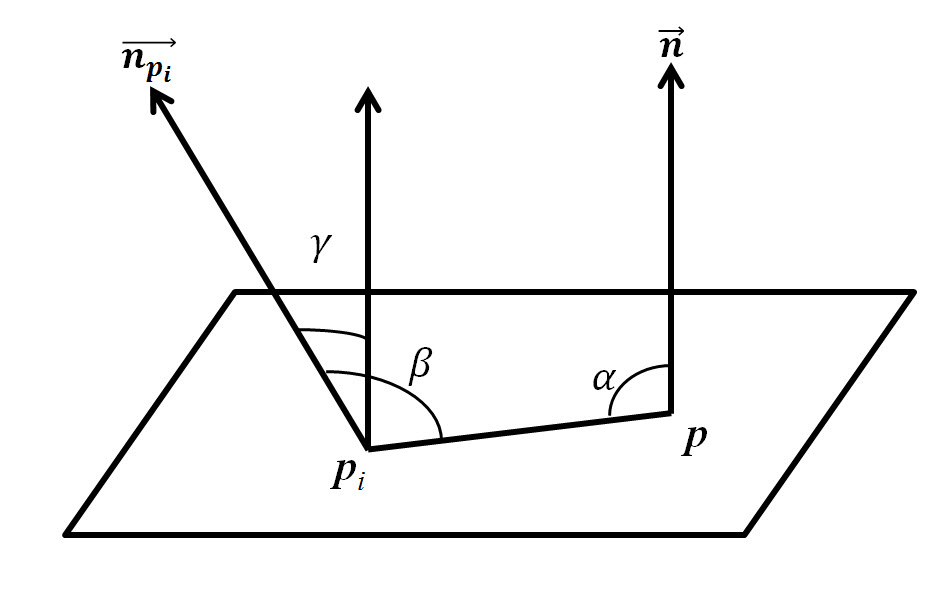}%
\label{fig_mcov}}
\hfil
\subfigure[]{\includegraphics[width=0.5in]{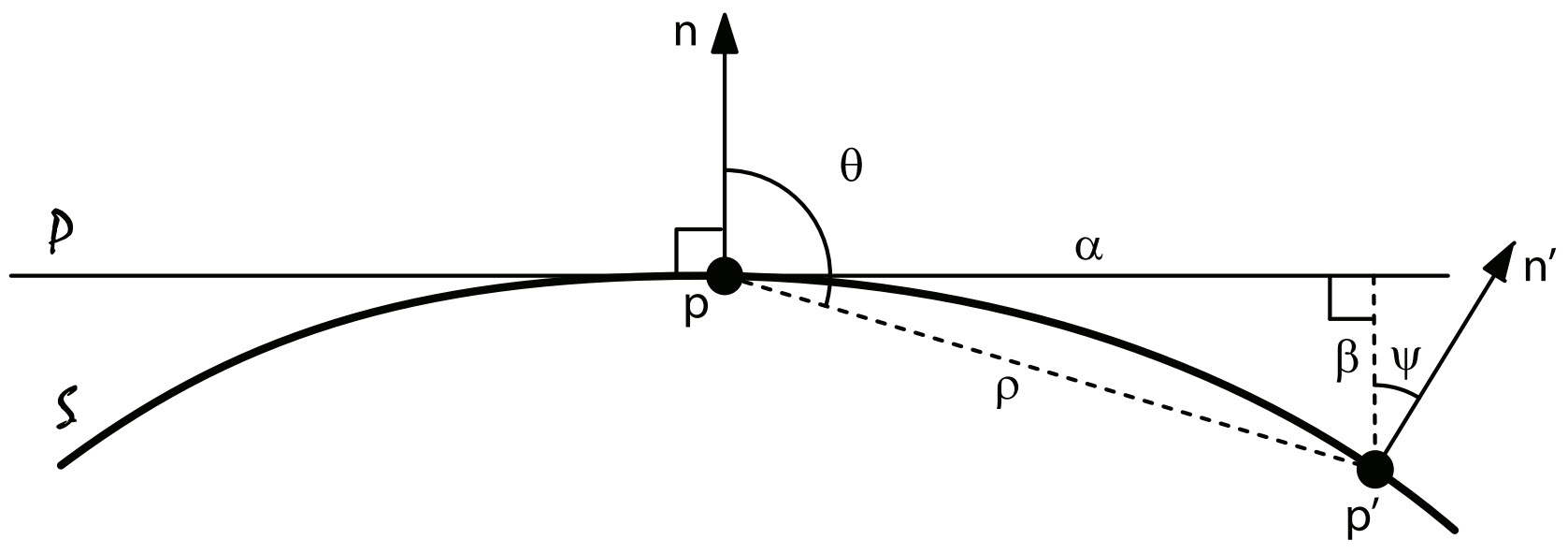}%
\label{fig_covariance}}
\hfil
\subfigure[]{\includegraphics[width=0.5in]{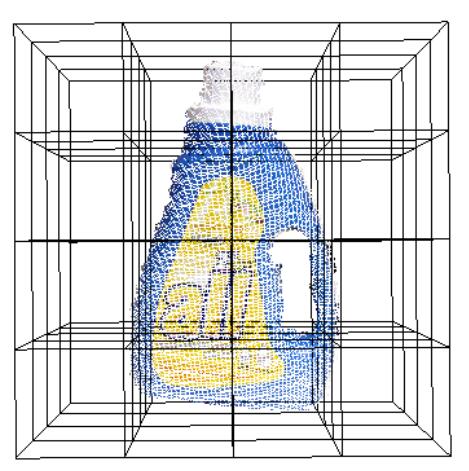}%
\label{fig_gasd}}
\caption{Different hand-crafted 3D point cloud descriptors. }
\label{fig_descriptors}
\end{figure*}

\subsection{Local-based Descriptor}
\label{Local-based Descriptor}
3D Local descriptors have been developed to encode the local information of feature points (e.g., surface normal and curvature), which is directly associated with the quality and resolution of the 3D point cloud model. \hl{Generally, these descriptors can be used in applications like .}

\subsubsection{Local reference frame based descriptor}
These methods commonly establish a local reference frame (LRF) computed for the feature point or key point. Then, the local neighborhood found for keypoint is partitioned into areas (e.g., bins) in accordance with the LRF. Finally, the descriptor is built by counting the spatial distribution quantities (named spatial distribution based method) or geometric attributes or relationships (called geometry-based method) in each area.    

\begin{itemize}
    \item [1)] \hl{Spatial distribution based method}
\end{itemize} 

     \textbf{Intrinsic Shape Signatures} is devised as following procedures. First, given a feature point \emph{\textbf{p}}, a LRF ($\emph{\textbf{e}}_{1}, \emph{\textbf{e}}_{2}, \emph{\textbf{e}}_{1}\bigotimes\emph{\textbf{e}}_{2}$) is computed, where \emph{\textbf{e}}$_{1}$, \emph{\textbf{e}}$_{2}$ and \emph{\textbf{e}}$_{3}$ are the eigenvectors obtained using the eigen analysis of \emph{\textbf{p}}'s spherical support. Second, the spherical angular space ($\theta,\mu$) constructed using octahedron is divided into several bins. And the final ISS descriptor is a 3D histogram created by counting the weighted sum of points in each bin. It is stated that the ISS is stable, repeatable, informative and discriminative. Ge et al. \cite{Ge2016Non} used a multilevel ISS strategy to extract descriptor to perform registration.
     
    \textbf{Normal aligned radial feature descriptor} \cite{Radu2010NARF} is estimated as following steps. A normally aligned range value patch around the feature point is computed by constructing a local coordinate system. Then a star-shaped pattern is projected into this patch to form the final descriptor. Furthermore, its rotationally invariant version is achieved by shifting the NARF according to a unique orientation extracted from the original counterpart. This approach obtains better results on feature matching.
    
    \textbf{Signature of Histogram of Orientation(SHOT)} \cite{Tombari2010Unique} can be considered as a combination of Signatures and Histograms. First, a repeatable LRF with disambiguation and uniqueness is computed for the feature point based on disambiguated Eigenvalue Decomposition of the covariance matrix of points within the support region. Then, the anisotropic spherical grid is used to define the signature structure that partitions the neighborhood along the radial, azimuth, and elevation axes (Figure \ref{fig_shot}). For each grid, point counts are accumulated into bins based on the angle between normals at feature point and its neighbors within this grid to obtain a local histogram. Moreover, the juxtaposition of these local histograms forms the final SHOT descriptor. In order to improve the accuracy of feature matching, Tombari et al. \cite{Tombari2011A} incorporated texture information (CIELab color space) to extend the SHOT to form its color version, i.e., SHOTCOLOR or CSHOT. Experimental results demonstrate that it presents a good balance between recognition accuracy and time complicity. Gomes et al. \cite{Beserra2013Efficient} introduced the SHOT descriptor on foveated point cloud to their objection recognition system to reduce the processing time. Furthermore, experiments show an attractive performance.
    Prakhya et al. \cite{Prakhya2015B} first employed a binary 3D feature descriptor, named \textbf{Binary SHOT (B-SHOT)} which is formed through replacing each value of SHOT with either 0 or 1. This construction procedure is performed by encoding every sequential four values taken from a SHOT descriptor, in turn into corresponding binary values according to the five possibilities defined by authors. This method requires significantly less memory and be obviously faster than SHOT.
    
    \textbf{Unique Shape Context} \cite{Tombari2010Uniqueshape} is an improvement of the 3D Shape descriptor by adding a unique, unambiguous LRF with the purpose of avoiding computation of multiple features at each keypoint. Given a query point \emph{\textbf{p}} and its spherical support region with radius \emph{R}, a weighted covariance matrix is defined as $ M=\frac{1}{Z}\sum_{i:d_{i}\leq R}(R-d_{i})(\textbf{p}_{i}-\textbf{p})(\textbf{p}_{i}-\textbf{p})^T$, where $Z=\sum_{i:d_{i}\leq R}(R-d_{i})$. Three unit vectors of LRF are computed from the Eigen Vector Decomposition of \emph{M}. The eigenvectors corresponding to the maximum and minimum eigenvalues are re-oriented in order to match the majority of the vectors they depicted, while the sign of the third eigenvector is determined by the cross product. The detailed definition of this LRF is presented in \cite{Tombari2010Uniqueorignal}. Once the LRF is built, the construction of the USC descriptor follows the approach analogous to that used in 3DSC. From the viewpoint of the memory cost and efficiency, the USC upgrades 3DSC notably.
    
    \textbf{Spectral Histogram} \cite{Behley2012} is similar to the SHOT descriptor. They first decomposed the covariance matrix defined in the support region to compute eigenvalues $\lambda_{0}$, $\lambda_{1}$, $\lambda_{2}$, where $\lambda_{0}$$\leq$$\lambda_{1}$$\leq$$\lambda_{2}$. The corresponding normalized eigenvalues are evaluated by$\lambda$$_{i}$$^{'}$=$\lambda$$_{i}$/$\lambda$$_{2}$. Then three signature values $\lambda$$_{0}$$^{'}$, $\lambda$$_{1}$$^{'}$-$\lambda$$_{2}$$^{'}$ and $\lambda$$_{2}$$^{'}$-$\lambda$$_{1}$$^{'}$ are calculated respectively. Finally, they subdivided these three values into different sectors (Figure \ref{fig_sh}) and accumulated the number of points falling into every sector to form the descriptor. This method is a best choice for the classification in urban environment. 
     
     \textbf{3D Self-similarity Descriptor} \cite{Huang2012} contains two major steps. The first step computes normal similarity $  s\left ( x,y,f_{normal} \right )=\left [ \pi - cos^{-1}\left ( \textbf{n}\left ( x \right ) -\textbf{n}\left ( y \right )\right ) \right ]/\pi$, curvature similarity $s\left ( x,y,f_{curv} \right )=1-\left | f_{curv}(x) - f_{curv}(y) \right |$ and photometric similarity $ s\left ( x,y,f_{photometry} \right )=1-\left | I(x)-I(y) \right |$, between two points \emph{x} and \emph{y}. Then, the united similarity is defined as $  s\left ( x,y\right )=\sum_{p\in PropertySet}w_{p}\cdot s\left ( x,y,f_{p} \right )$. By comparing the feature point's united similarity with that of its neighbors, the self-similarity surface is directly constructed. The second step is to build a LRF at the feature point to guarantee the rotation invariance and quantize the correlation space into cells with average similarity value of points falling in corresponding cell to form the descriptor (Figure \ref{fig_3DSSIM}). This descriptor can efficiently characterize the distinctive geometric signatures in point cloud.
    
    Tang et al. \cite{Tang2016Signature} generated a new \textbf{ Signature of Geometric Centroids (SGC)} descriptor by first constructing a unique LRF centered at feature point \emph{\textbf{p}} based on PCA. Then its local spherical support \emph{S$_{p}$} aligns with the LRF and a cubical volume encompassing \emph{S$_{p}$} is defined with edges length of 2\emph{R} following by partitioning this volume evenly into \emph{K}$\times$\emph{K}$\times$\emph{K} voxels. Finally, the descriptor is constructed by concatenating the number of points within each voxel and the position of centroid calculated for these points. The dimension of SGC descriptor is 2$\times$\emph{K}$\times$\emph{K}$\times$\emph{K}.
    
    \textbf{3D Histogram of Point Distribution descriptor} \cite{Prakhya20173DHoPD} are formed using the following steps. First, the local surface of keypoint is aligned with the defined LRF to achieve rotational invariance. Next, along the x-axis, the histograms are made by portioning the range between the smallest and largest x coordinate value of the points in the surface into D bins and accumulating the points falling in each bin. Repeating the same processing along y and z axis, the 3DHoPD is generated by concatenating these histograms together. The advantage of this method is that the computation is greatly fast.

\begin{itemize}
    \item [2)] \hl{Geometry based method}
\end{itemize} 

    \textbf{Point Feature Histogram, PFH} \cite{Rusu2008}\cite{Rusu2008Aligning}, uses the relationships between point pairs. For every pair of points \textbf{\emph{p$_{i}$}} and \textbf{\emph{p$_{j}$}} in the neighborhood of \textbf{\emph{p}} (Figure \ref{fig_pfhdiagram}), where one point is chosen as \textbf{\emph{p$_{s}$} }and the other as \textbf{\emph{p$_{t}$}}. First, a Darboux frame is constructed at \textbf{\emph{p$_{s}$}} as (Figure \ref{fig_pfhframe}): $\textbf{u }= \textbf{n}_{s}, \textbf{v} =\textbf{u} \times \frac{\left ( \textbf{p}_{t} - \textbf{p}_{s} \right )}{\left \| \textbf{p}_{t} - \textbf{p}_{s} \right \|_{2}}, \textbf{w} = \textbf{u} \times \textbf{v}$. Then, using this frame, three angular features, $  \alpha =\left \langle \textbf{v},\textbf{n}_{t} \right \rangle, \Phi = \left \langle \textbf{u},\textbf{p}_{t} - \textbf{p}_{s} \right \rangle / d, \theta = arctan(\left \langle \textbf{w},\textbf{n}_{t} \right \rangle,\left \langle \textbf{u},\textbf{n}_{t} \right \rangle)$, expressing the difference between normals \textbf{\emph{n$_{t}$}} and \textbf{\emph{n$_{s}$}}, and the distance $d = \left \| \textbf{p}_{t}-\textbf{p}_{s} \right \|_{2}$ are computed. The final PFH representation is created by binning these four features into a histogram with \emph{div$^{4}$} bins, where \emph{div} is the number of subdivisions along each features' value range.
    
    However, higher computational complexity makes PFH inappropriate to be used in real-time applications. Therefore, \textbf{Fast Point Feature Histogram (FPFH)} \cite{Rusu2009Fast}\cite{Rusu2009Detecting} is proposed to solve this problem, which consists of two steps. The first step is the construction of the Simplified Point Feature Histogram (SPFH). Three angular features $\alpha$, $\Phi$, $\theta$ between query point and its neighbors are computed using the same way as PFH, which are binned into three separate histograms. In the second step, the final FPFH of the query point \emph{p$_{q}$} is calculated as $FPFH\left ( p_{q} \right ) = SPFH\left (  p_{q} \right ) + \frac{1}{k}\sum_{i=1}^{k}w_{k} \cdot SPFH\left (  p_{k} \right )$, where \emph{w$_{k}$} donates the distance between the query point and its neighbor (Figure \ref{fig_fpfhframe}). The FPFH descriptor can greatly reduce the computational complexity to \emph{O}(\emph{nk}). Huang et al. \cite{Huang2013} combined the FPFH descriptor and SVM (Support Vector Machine) learning algorithms for detecting objects in Scene Point Cloud, achieving effective results for cluttered industrial scenes.
\begin{itemize}
    \item [3)] \hl{Combination of geometry and color based method}
\end{itemize}
    
    Similar to PFH and FPFH descriptors, \textbf{ Colored Histogram of Spatial Concentric Surflet-Pairs (CoSPAIR)} descriptor in \cite{Logoglu2016CoSPAIR} also relies on surflet-pair relations \cite{Wahl2003Surflet}. Considering the query point \emph{\textbf{p}} and its neighbor \emph{\textbf{q}} within the spherical support, a fixed LRF and three angles between them are estimated using the same strategies as PFH. Then, the neighboring space is partitioned into several spherical shells (called level) along the radius. For each level, three distinct histograms are generated by accumulating points in it along with the three angular features. The original SPAIR is the concatenation of the histograms at all levels. Finally, CIELab color space is binned into histograms for each channel at each level, together with the SPAIR descriptor, to form the final CoSPAIR descriptor. It is reported that this method simple and fast.
    

\subsubsection{Non-local reference frame based descriptor}
This type of descriptor establishes the spatial or geometric relationships between each feature point (keypoint) and its neighbors to depict the local information without certain LRF.

\begin{itemize}
    \item [1)] \hl{Signature based approach }
\end{itemize}
    \hl{This type of approach encodes the geometric information computed around the local support region of each feature point to reflect the feature of the local neighborhood.}
   
   \textbf{Spin Image} \cite{Johnson1998Surface}: The spin image attributes of each neighboring point \emph{\textbf{q}} of the feature point \emph{\textbf{p}} is defined as a pair of distances ($\alpha$,$\beta$), where $\alpha = \mathbf{n}_{q}\cdot \left ( \mathbf{p}-\mathbf{q} \right )$ and $\beta = \sqrt{\left \| \textbf{p}-\textbf{q} \right \|^{2}-\alpha ^{2}}$ (shown in Figure \ref{fig_SI}). And the final descriptor is generated by accumulating the neighbors of feature point in discrete 2D bins. This descriptor is robust to occlusion and clutter \cite{765655}, but the presence of high-level noise would lead to degradation of performance.

  Matei et al. \cite{Matei2006} incorporated the spin image descriptor into their work to handle the recognition of 3D point cloud of vehicles, which is a challenging problem. Similarly, Shan et al. \cite{Shan2006} also took the spin image as descriptive descriptors to propose the Shapeme Histogram projection(SHP) approach, together with a Bayesian framework to complete partial object recognition through the projection of the descriptor of the query model onto the subspace of the model database. Golovinskiy et al. \cite{Golovinskiy2009} applied the shape-based spin image descriptor in conjunction with contextual features to form a discriminative descriptor to identify the urban objects. 
  
  \textbf{Eigenvalues based Descriptors} \cite{Vandapel2004} is used to extracted saliency features. The eigenvalues are attained by decomposing the local covariance matrix defined in a local support region around feature points and decreasingly ordered $\lambda_{0}\geq\lambda_{1}\geq\lambda_{2}$. Three saliencies, named point-ness $\lambda _{2}$, curve-ness $\lambda _{0}-\lambda _{1}$ and surface-ness $\lambda _{1}-\lambda _{2}$ respectively, are constructed by means of the linear combination of these eigenvalues. In literature  \cite{Anand2013Contextually}\cite{Kahler2013Efficient}\cite{Zelener2015Classification}, the above three features also are integrated into the corresponding feature representations to perform object detection in indoor scenes, 3D scene labeling and vehicles detection, respectively.
  
  \textbf{Radius-based Surface Descriptor (RSD)} \cite{Marton2010General} depicts the geometric property of point by estimating the radial relation with its neighbouring points. The radius is modeled as relation between distance of two points and the angle between their normals as follows$  d_{\alpha }=\sqrt{2}r\sqrt{1-cos(\alpha )}=r\alpha + r\alpha ^{3}/24+O(\alpha ^{5})$. By solving this equation, the maximum radius and minimum radius are obtained to constructed the final descriptor for each point as \emph{\textbf{d}}$_{i}$ = [\emph{r}$_{max}$,\emph{r}$_{min}$]. The advantage of this method is simple and descriptive.
  
  \textbf{Depth kernel Descriptor} \cite{Bo2011Depth} extends kernel descriptors to 3D point cloud to derive five local kernel descriptors that describe size, shape, and edges, respectively. Experimental results show that these features can complement each other, and this method turns out to significantly improve the accuracy of object recognition. 
  
  \textbf{Covariance based descriptor} \cite{Fehr2012Compact} is exploited for 3D point cloud due to the compactness and flexibility of containing multiple features for representational power. These features they decided to capture the geometric relation include $\alpha$,$\beta$,$\theta$,$\rho$,$\psi$ and \emph{n}$^{'}$ (Figure \ref{fig_covariance}). The advantages of this method are low computational and storage requirements, no model parameters to tune, and scalability relevant to various features. To further increase performance, Fehr et al. \cite{Fehr2014RGB} took the additional r,g,b color channel values from 'colored' point cloud into consideration to form a straightforward extension of the original covariance-based descriptors. And this approach actually yields promising results. Beksi et al. \cite{7139443} computed covariance descriptors that also encapsulated both shape features and visual features on the entire point cloud. However, the difference from \cite{Fehr2014RGB} is that they incorporated the principal curvatures and Gaussian curvature into shape vector while adding gradient and depth into the visual vector. Then, together with dictionary learning, a point cloud classification framework has been constructed to classify the object.
  
\textbf{MCOV} descriptor \cite{Cirujeda2014MCOV} fuses visual and 3D shape information. Given a feature point \emph{\textbf{p}} and its radial neighborhood \emph{N}$_{p}$, a feature selection function is defined as $  \Phi (\emph{\textbf{p}},N_{p}) = \left \{\varnothing _{p_{i}},\emph{\textbf{p}}_{i}\in N_{p}  \right \}$, where $\varnothing$$_{p_{i}}$ is represented as a vector (R$_{p_{i}}$,G$_{p_{i}}$,B$_{p_{i}}$,$\alpha$$_{p_{i}}$,$\beta$$_{p_{i}}$,$\gamma$$_{p_{i}}$) (as shown in Figure \ref{fig_mcov}). The first three elements correspond to the R, G, B values at \emph{\textbf{p$_{i}$}} in RGB space capturing the texture information. And the last three components are $\langle$\emph{\textbf{p}},(\emph{\textbf{p$_{i}$}}-\emph{\textbf{p}})$\rangle$, $\langle$\emph{\textbf{p$_{i}$}},(\emph{\textbf{p$_{i}$}}-\emph{\textbf{p}})$\rangle$ and $\langle$\emph{\textbf{n$_{p}$}},\emph{\textbf{n$_{p_{i}}$}}$\rangle$, respectively. Then, a covariance descriptor at \emph{\textbf{p}} is calculated by $  C_{r}(\Phi (\emph{\textbf{p}},N_{p}))=\frac{1}{N-1}\sum_{i=1}^{N}(\varnothing _{p_{i}}-\mu )(\varnothing _{p_{i}}-\mu )^{T}$, where $\mu$ is the mean of the $\varnothing_{p_{i}}$. Results on testing point clouds demonstrate that the MCOV providing a compact and depictive representation boosts the discriminative performance dramatically.
 \begin{itemize}
         \item [2)] \hl{Single-type statistic information Histogram based approaches }
 \end{itemize}   
\hl{These methods describe the local feature in the form of the histogram by summing up simply the spatially or geometrically statistic information in each region (e.g., cube, grid) formed by dividing the neighboring support region of each feature point according to certain strategies (e.g. Orientation partition ).  }
    
    \textbf{3D shape contexts(3DSC)} \cite{Frome2004} directly extends the 2D shape context descriptor to 3D point cloud. A spherical support region, centered on a given feature point \emph{\textbf{p}}, is first determined with its north pole orienting as the surface normal \emph{\textbf{n}}. Within the support, a set of bins (Figure \ref{fig_3DSC}) is formed by equally dividing the azimuth and elevation and logarithmically spacing the radial dimension. Then, the final 3DSC descriptor is computed as the weighted sum of the number of points falling into bins. Actually, this descriptor captures the local shape of point cloud at \emph{\textbf{p}} using the distribution of points in a spherical support. However, it requires the computation of multiple descriptors for each feature point because of no definition of Reference Frame at the feature point.
    
    \textbf{Histogram of Normal Orientation} \cite{Triebel2006} represents a local distribution of the cosine of the angles between the surface normal on \emph{\textbf{p}} and the normals of its neighbors, which is then defined as a local histogram. In general, regions with a strong curvature result in a uniformly distributed histogram, while flat areas lead to a peaked histogram \cite{Behley2012}. 
    
    \textbf{ThrIFT} \cite{Flint2007Thrift} takes orientation information into account. For each keypoint \emph{\textbf{p}} and point \emph{\textbf{q}} from its neighbouring support, two windows \emph{W}$_{1}$ and \emph{W}$_{2}$ are computed before estimating their least-squares plane's normals \emph{\textbf{n}}$_{small}$ and \emph{\textbf{n}}$_{large}$ (shown in Figure \ref{fig_Thrift}). The output descriptor for \emph{\textbf{p}} is generated by binning the angle $\theta$ between \emph{\textbf{n}}$_{small}$ and \emph{\textbf{n}}$_{large}$ into a histogram, $cos(\theta )= \frac{\textbf{n}_{small}\cdot \textbf{n}_{large}}{\left \| \textbf{n}_{small} \right \|\left \| \textbf{n}_{large} \right \|}$. Although the ThiIFT descriptor can yield promising results, it is sensitive to noise.
    
    To handle viewpoint variations and effect of noise, Rahmani et al. \cite{Rahmani2014HOPC} proposed a 3D point cloud descriptor, named \textbf{Histogram of Oriented Principal Components}(HOPC). First, PCA is performed on the support centered keypoint to yield the eigenvectors and corresponding eigenvalues. Then, they projected each eigenvector into \emph{m} directions derived from a \emph{regular m-sided polyhedron} and scaled it by the corresponding eigenvalue. Finally, the projected eigenvectors in decreasing order of eigenvalues are concatenated to form the HOPC descriptor.
     
     \textbf{Height Gradient Histogram (HGIH)}, developed by Zhao et al. \cite{Zhao2014Height}, takes full advantage of the height dimension data which is firstly extracted from 3D point cloud as \emph{f}(\emph{\textbf{p}})=\emph{p$_{x}$} (\emph{x} corresponds to height) for each point \emph{\textbf{p}}=(p$_{x}$,p$_{y}$,p$_{z}$). After that, the linear gradient reconstruction method is used to compute the height gradient $\nabla$\emph{f}(\emph{\textbf{p}}) of point \emph{\textbf{p}} based on its neighbors. Secondly, the spherical support of each point \emph{\textbf{p}} is divided into \emph{K} sub-regions and the gradient orientation of points with one sub-region is encoded to form a histogram. Finally, the HIGH feature descriptor is constructed by concatenating the histograms of all sub-regions. Experimental results show that the HIGH descriptor can give promising performance. However, one major limitation is its lack of capability of describing small objects well.
     
     \textbf{Local Feature Statistics Histograms}: Yang et al. \cite{Yang2016A} exploited the statistical properties of three local shape geometry: local depth, point density and angles between normals. Given a keypoint \emph{\textbf{p}} and its spherical neighborhood \emph{\textbf{N}}, all neighbors \emph{\textbf{p$_{i}$}} are projected on a tangent plane estimated along the normal \emph{\textbf{n}} at \emph{\textbf{p}} to form new points \emph{\textbf{p$_{i}$$^{'}$}}. The local depth is then computed as \emph{d}=\emph{r}- \emph{\textbf{n}}$\cdot$(\emph{\textbf{p$_{i}$$^{'}$}}-\emph{\textbf{p$_{i}$}}). And the angular feature is denoted as $\theta$=arccos(\emph{\textbf{n}},\emph{\textbf{n$_{p_{i}}$}}). Regarding point density, the ratio of points falling into each annulus generated by equally dividing the circle on the plane is determined via horizontal projection distance $\rho =\sqrt{\left \| \mathbf{p}^{'}-\mathbf{p}_{i}^{'} \right \|-(\mathbf{n}\cdot (\mathbf{p}^{'}-\mathbf{p}_{i}^{'}))^{2}}$. The final descriptor is constructed using the concatenation of the histograms built on these three features, which has low dimension, low computational complexity and is robust to various nuisances.
\begin{itemize}
         \item [3)] \hl{Multi-type statistic information Histogram based approaches }
\end{itemize}
 \hl{These are a kind of method that takes advantage of multiple types of statistic information (spatial distribution, color or intensity, and geometric attributes) of the local neighborhood of each feature point to formulate individually histogram. The final descriptor is the concatenation of these histograms. }
     
     \textbf{Rotation, illumination, scale invariant appearance and shape feature (RISAS)}, presented by Li et al. \cite{Li2016RISAS}, is a feature vector including three statistical histograms, namely spatial distribution, intensity information and geometrical information. As for spatial distribution, a spherical support is divide into \emph{n$_{pie}$} sectors and information (position and depth value) in each sector is encoded to form spatial histogram. Regarding intensity information, relative intensity instead of absolute intensity is grouped into \emph{n$_{bin}$} bins to construct the intensity histogram. Finally, with respect to geometric information, the quantity $ \rho _{i} = \left | \left \langle  \textbf{n}_{p}, \textbf{n}_{q} \right \rangle \right |$ between normal \textbf{n$_{p}$} at feature point \emph{\textbf{p}} and normal \textbf{n$_{q}$} at its neighbor \emph{\textbf{q}} is computed and these values are then binned to form the geometric histogram. Experimental results show the effectiveness of proposed descriptor in point cloud alignment.

\subsection{Global-based Descriptor}
Global descriptors encode the geometric information of the whole 3D point cloud, which requires relatively less computation time as well as memory footprint. \hl{In general, global descriptors are increasingly used for 3D object recognition, geometric categorization, and shape retrieval.}

\subsubsection{Geometry based method}
This kind of method makes full use of geometric properties or relationships between points to describe the characteristic of the entire point cloud, mostly in the form of a histogram. 

Similar to surflet-pair feature, given two points \emph{\textbf{p}}$_{1}$ and \emph{\textbf{p}}$_{2}$ and their normals \emph{\textbf{n}}$_{1}$ and \emph{\textbf{n}}$_{2}$, the \textbf{Point Pair Feature} (PPF) \cite{Drost2010Model} is defined as $F=(\|\emph{\textbf{p}}_{1}-\emph{\textbf{p}}_{2}\|_{2}, \angle(\emph{\textbf{n}}_{1},\emph{\textbf{p}}_{2}-\emph{\textbf{p}}_{1}), \angle(\emph{\textbf{n}}_{2},\emph{\textbf{p}}_{2}-\emph{\textbf{p}}_{1}),\angle(\emph{\textbf{n}}_{1},\emph{\textbf{n}}_{2}))$, where the range of angle is [0;$\pi$]. Feature with the same discrete version are aggregated together. And the global descriptor is formed by mapping the sampled PPF space to the model.

\textbf{Global RSD} \cite{Marton2011Combined} can be regarded as a global version of the local RSD descriptor. After voxelizing the input point cloud, the smallest and largest radius is estimated in each voxel, which is labeled using one of the five surface primitives (e.g., planar, sphere) by checking the radii. Once all voxels have bee labeled, a global RSD descriptor can be designed by analyzing the relations between all labels.

\textbf{Viewpoint Feature Histogram (VFH)} \cite{Muja2011REIN}\cite{Rusu2014Fast} extends the idea and properties of FPFH by including additional viewpoint variance. It is a global descriptor composed of a viewpoint direction component and a surface shape component. The viewpoint component is a histogram of angle $\beta =arccos(\textbf{n}_{p}\cdot (\textbf{v}-\textbf{p})/\left \| \textbf{v}-\textbf{p} \right \|_{2})$ between central viewpoint direction and each point's surface normal. As for shape component, three-angle $\alpha$, $\phi$, and $\theta$ computed similarly as PFPF are binned into three distinct sub-histograms, respectively, each with 45 bins. The VFH turns out to have high recognition performance and computational complexity of \emph{O}(\emph{n}). However, the main drawback of VFH is its sensitivity to noise and occlusions. Ali et al. \cite{Ali2014Contextual} integrated the VFH descriptor into their proposed system to perform scene labeling. Chan et al. \cite{Chan2014A} extracted the VFH feature from a point cloud of human to calculate the human-pose estimation.

\textbf{The Clustered Viewpoint Feature Histogram descriptor} \cite{Aldoma2012CAD} can be considered as an extension of VFH, which takes into account the advantage from stable object regions obtained by applying a region growing algorithm after removing the points with high curvature. Given a region \emph{s}$_{i}$ from the regions set \emph{S}, a Darboux frame (\textbf{\emph{u}}$_{i}$,\textbf{\emph{v}}$_{i}$,\textbf{\emph{w}}$_{i}$) similar to FPH is constructed using the centroid \emph{\textbf{p}}$_{c}$ and its corresponding normal \emph{\textbf{n}}$_{c}$ of \emph{s}$_{i}$. Then the angular information ($\alpha,\phi,\theta,\beta$) as in VFH are binned into four histograms followed by computation of Shape Distribution Component (SDC) of CVFH. $SDC=\frac{(\textbf{p}_{c}-\textbf{p}_{i})^{2}}{max((\textbf{p}_{c}-\textbf{p}_{i})^{2})}$. The final CVFH descriptor is the concatenation of histograms created from ($\alpha,\phi,SDC,\theta,\beta$) with the size of 308. Although the CVFH can produce promising results, lacking of notion of an aligned Euclidean space causes the feature to miss a proper spatial description \cite{Aldoma2012OUR}.

\textbf{The Oriented, Unique, and Repeatable Clustered Viewpoint Feature Histogram (OUR-CVFH)} has the same surface shape components and viewpoint component as CVFH, but the dimension of viewpoint component is reduced to 64 bins. On the other hand, eight histograms of distances between points of surface \emph{S} and centroid are built to replace the shape distribution component in CVFH based on reference frames that are yielded by employing the Semi-Global Unique Reference Frame (SGURF) method. The size of the final OUR-CVFH is 303.

\subsubsection{Spatial distribution based method}
These methods firstly partition the 3D point cloud into several grids or cells or special bins according to designed rules (e.g., building a global reference frame). Then, the feature descriptor is formed by accumulating the spatial distribution arising in each division unit aforementioned.

\textbf{ Global Structure Histograms (GSH)} \cite{Madry2012Improving} descriptor encodes the global and structural properties in the 3D point cloud in the three-stage pipeline. Consider a point cloud model, first, a local descriptor is estimated for each point, and then all points based on their descriptors are labeled one approximated surface class by using k-means algorithm followed by the computation of Bag-of-Words model. In the second step, the relationship between different classes is determined along with the surface form by triangulation. The final step is the construction of the GSH descriptor representing the object as the distribution of distance along the surface in the form of a histogram. This descriptor not only maintains the low variations but also reflects global expressiveness power.

Aiming at real-time application, Wohlkinger et al. \cite{Wohlkinger2012Ensemble} introduced a global descriptor from the partial point cloud, dubbed \textbf{Ensemble of Shape Functions} which is a concatenation of ten 64-sized histograms of shape distributions including three-angle histograms, three area histograms, three distance histograms, and one distance ratio histogram. The first nine histograms are created by respectively classifying the A3 (angle formed by randomly sampled three points), D3 (area created by three points) and D2 shape function (line between point-pair sampled from point cloud) into ON, OFF and MIXED cases according to the mechanism mentioned in \cite{Ip2002Using}, while the distance ration histogram is built on the lines from D2 shape function. The final ESF descriptor has proven to be efficient and expressive.

\textbf{The Global Fourier Histogram (GFH)} descriptor \cite{Chen2014Performance} is generated for an oriented point which is chosen as the original point \emph{\textbf{0}}. The normal at \emph{\textbf{0}} together with global z-axis (\emph{\textbf{z}}=[0,0,1]$^T$) are utilised to construct the global reference frame. Whereafter, a 3D cylindrical support region centered at \emph{\textbf{0}} is divided into several bins by equally spacing elevation, azimuth, and radial dimensions. The GFH descriptor is a 3D histogram formed by summing up the points in each bin. To further boost the robustness, 1D Fast Fourier Transform is applied to analyze the 3D histogram along the azimuth dimension. This descriptor compensates for the drawback of the Spin Image method.

In order to improve vehicle detection rate remarkably, three novel features in \cite{Cheng2014Robust}, namely position-related shape, object height along the length, and reflective intensity histogram, are extracted together to distinguish vehicles and other objects robustly. The position-related shape takes both shape features (including width to length ratio and width to height ratio) and position information (e.g., distance to the object, angle of view and orientation) into account to address the variance of orientation and angle of view. Furthermore, in the next stage, the bounding box around the vehicles is divided into several blocks along the length, and the average height in each block is added in the feature vector to improve discrimination further. Finally, a reflective intensity histogram with 25 bins is computed using the characteristic intensity distribution of vehicles. The experimental results show that the final feature is contributed greatly to classification performance.

\textbf{The Global Orthographic Object Descriptor (GOOD)} \cite{Kasaei2016GOOD} is constructed by first defining a unique and repeatable Local Reference Frame based on the Principal Component Analysis. Then, the point cloud model is orthographically projected onto three plans, called \emph{XoZ, XoY, and YoZ}. As for \emph{XoZ}, this plane is divided into several bins, and distribution matrices are computed by accumulating points falling into each bin. Moreover, similar processing is performed for \emph{XoY} and \emph{YoZ}. Afterward, two statistical features, i.e. entropy and variance, are estimated for each distribution vector transformed from the matrix above. Concatenating these vectors together forms a single vector for the entire object. Experimental results report that the GOOD is the scale and pose invariant and can give a trade-off between expressiveness and computational cost.

\textbf{Globally Aligned Spatial Distribution (GASD)}, a novel global descriptor proposed by Lima et al. \cite{Lima2016}, mainly consists of two steps. A reference frame estimated for the entire point cloud model is constructed based on PCA, where x and z-axis are the eigenvectors \emph{\textbf{v$_{1}$}},\emph{\textbf{v$_{3}$}} corresponding to the minimal and maximal eigenvalues, respectively, of covariance matrix \emph{\textbf{C}} that is computed from \emph{\textbf{P$_{i}$}} and the centroid $\overline{\emph{\textbf{P}}}$ of point cloud, y-axis is \emph{\textbf{v$_{2}$}}=\emph{\textbf{v$_{1}$}}$\times$\emph{\textbf{v$_{3}$}}. Then, the whole point cloud is aligned with this reference frame. Next, a axis-aligned bounding cube of point cloud-centered at $\overline{\emph{\textbf{P}}}$ is partitioned into \emph{m$_{s}$}$\times$\emph{m$_{s}$}$\times$\emph{m$_{s}$} cells (Shown in Figure \ref{fig_gasd}). The global descriptor is achieved by concatenating the histograms formed by summing up the number of points falling into each grid. To get a higher discriminative power, color information based on HSV space is incorporated into the descriptor in a similar way as computing shape descriptor to form the final descriptor. However, this method may not work well with objects having similar shape and color distribution.

\textbf{Scale Invariant point Feature (SIPF)} \cite{Lin2017S} computes \emph{\textbf{q}}$^{*}$=argmin$_{q}$ $\|$\emph{\textbf{p}}-\emph{\textbf{q}}$\|$ between feature point \emph{\textbf{p}} and border point \emph{\textbf{q}} as the reference direction. Then, the angle of a local cylindrical coordinates with \emph{\textbf{q}}$^{*}$ is partitioned into several cells. The final SIPF descriptor is constructed by concatenating all the normalized cell features \emph{D}$_{i}$=\emph{exp}(\emph{d}$_i$/(1-\emph{d})), where \emph{d}$_{i}$ is the minimum distance between \emph{\textbf{p}} and points in
\emph{i}th cell.

\subsection{Hybrid-based Descriptor}
 The hybrid-based descriptor is the sort of descriptor fusing the essential theorem of local and global descriptors or incorporating both kinds of descriptors together to make the most of the advantages of local and global features.
 
\textbf{Bottom-Up and Top-Down Descriptor:} Alexander et al. \cite{Alexander2008Object} combined bottom-up and top-down descriptors to operate on the 3D point cloud. In the Bottom-Up stage, spin images are computed for point cloud followed by the Top-Down stage, in which global descriptor-Extended Gaussian Images(EGIs)-are used to capture larger-scale structure information to depict the models further. Experimental results demonstrate that this approach provides balance
between efficiency and accuracy.

\textbf{Local and Global Point Feature Histogram:} For real-time application, a novel feature descriptor resorting to local and global properties of point cloud is described by Himmelsbach et al. \cite{Himmelsbach2009Real} in detail. First, four object-level features, including maximum object intensity, object intensity
mean and variance, and volume, are estimated as the global part. Then, as for the description of local point properties, three histograms are built for three features each, i.e. scatter-ness, linear-ness, and surface-ness from Lalonde et al. \cite{Lalonde2006Natural} followed by the adoption of Anguelove et al.'s  \cite{Anguelov2005} feature extraction approach, which produces five more histograms. The finally designed descriptor is proved well suited for object detection in large-sized 3D point cloud scenes. Lehtom\"{a}ki et al. \cite{Lehtom2016Object} integrated the LGPFH descriptor above
into their workflow to recognize objects (such as tree, car, pedestrian, lamp, etc.) in road and street environment.

\textbf{Local-to-Global Signature Descriptor:} To overcome the drawbacks of both local and global descriptors, Hadji et al. \cite{Hadji2014Local} proposed Local-to-Global descriptor (LGS). First, they classified the points based on the minimum radius of RSD \cite{Marton2010General} to capture the whole geometric property of the object. Next, both local and global support regions aggregated by the same class are adopted to describe feature points (local description). Finally, the LGS is created in a signature form. Since this descriptor makes use of the strengths of both local and global methods, so it is highly discriminative.

\textbf{Point-Based Descriptor:} Wang et al. \cite{Wang2015A} extracted point-based descriptor from point cloud as the foundation of their multiscale and
hierarchical classification framework. The first part of feature (denoted as \textbf{F}$_{eigen}$) is obtained using eigenvalues (\emph{$\lambda$}$_{1}$,\emph{$\lambda$}$_{2}$,\emph{$\lambda$}$_{3}$, in decreasing order) computed for covariance matric of feature point. $ \textbf{F}_{eigen}=\left [ \sqrt[3]{\coprod_{i=1}^{3}\lambda _{i}} ,\frac{\lambda _{1}- \lambda _{3}}{\lambda _{1}},\frac{\lambda _{2}- \lambda _{3}}{\lambda _{1}},\frac{\lambda _{3}}{\lambda _{1}}-\sum_{i=1}^{3}\lambda _{i}log(\lambda _{i}), \frac{\lambda _{1}- \lambda _{2}}{\lambda _{1}},\right] $. Spin image is adopted as the second part of their wanted feature, represented as \textbf{F}$_{SI}$. The final descriptor is the combination of \textbf{F}$_{eigen}$ and \textbf{F}$_{SI}$ which is an 18 dimensional vector.

\subsection{\hl{Summary of Hand-crafted Descriptor}}

Table 1 summarizes the characteristics of local-based, global-based, and hybrid-based descriptors extraction approaches. These approaches are arranged in chronological order.

\begin{table*}[h]
\caption{"Noise": robust to noise, "Density": robust to varying density, "Occlu. and Clut.": robust to occlusion and clutter, "P": Pose, "R": rotation, "S": Scale, "T": Translation, "V": Viewpoint, "I": Illumination,“O”: Orientation change,"SUOD": Sydney Urban Objects Datasets, “WD”: Wachtberg dataset, "WOD":Washington RGB-D Object, "BD": BigBIRD Dataset, "CD": Challenge Dataset, "PSB": Princeton Shape Benchmark, "SOC": Stereo Object Category.}
\label{tab:handcraft}
\resizebox{\textwidth}{!}{
\begin{tabular}{|c|c|c|c|c|>{\columncolor{yellow}}c|>{\columncolor{yellow}}c|>{\columncolor{yellow}}c|>{\columncolor{yellow}}c|>{\columncolor{yellow}}l|}
\hline
\multirow{2}*{No.}& \multirow{2}*{Name} & \multirow{2}*{Year} & \multirow{2}*{Size} & \multirow{2}*{Application} & \multirow{2}*{Noise} & \multirow{2}*{Density} & \multirow{2}*{Occlu.} & \multirow{2}*{Invariant} &\multirow{2}*{Performance}\\
~ & & & & & &  & \& clut.  & \\
\hline
1 & SI \cite{Johnson1998Surface} & 1998 & 225 & Recognition & - & - &  Y & P & 36.7\% on WOD.\\
2 & 3DSC \cite{Frome2004} & 2004 & 1980 & Recognition & Y& - & Y & R &21.7\% Outperforms SI on cluttered scenes. \\
3 & Thrift \cite{Flint2007Thrift} & 2007 & 32 & Recognition & - & - & - & R,V  & Be robust to data missing. \\
4 & BUTD \cite{Alexander2008Object} & 2008 & - & Detection & Y & - & Y &  R & Detects vehicle in large-scale scenario.\\
5 & PFH \cite{Rusu2008Aligning} & 2008 & 125 & Registration & Y & Y & - & P, R, T & Copes with data from indoor and outdoor laser scans. \\
6 & LGPFH \cite{Himmelsbach2009Real} & 2009 & 60 & - & - & - & - & - &Performs real time object classification. \\
7 & FPFH \cite{Rusu2009Fast} & 2009 & 33 & Registration & Y & Y & - & S & Be suitable for real time application.\\
8 & GFPFH \cite{Rusu2009Detecting} & 2009 &16 & Recognition & - & - & Y & - & 43.3\% on WOD. \\
9 & ISS \cite{Zhong2010Intrinsic} & 2010 & Vary & Recognition & Y & - & Y & - & Outperforms SI, 3DSC in the presence of noise, occlusion and clutter. \\
10 & PPF \cite{Drost2010Model} &  2010 & - & Recognition & Y & - & Y & R,T & Allows for sparser point cloud model and obtain fast performance.\\
11 & SHOT \cite{Tombari2010Unique}& 2010 & 352 & Recognition & Y & - & Y & R, S & Outperforms SI on Spacetime Stereo dataset. \\
12 & USC \cite{Tombari2010Uniqueshape}& 2010 &1980 & Recognition & Y & - & Y & R, T &38.3\% on WOD and outperforms 3DSC in terms of memory cost and efficiency. \\
13 & Kernel \cite{Bo2011Depth} & 2011 & Vary & Recognition & - & - & - & - & 78.8\% on WOD.\\
14 & GRSD \cite{Marton2011Combined} & 2011 & 20 & - & - & -& & S &   46.0\% on WOD.\\
15 & VFH \cite{Muja2011REIN} & 2011 &308 & Recognition & - & - & - & R & 76.7\% on WOD and outperforms SI.  \\
16 & CSHOT \cite{Tombari2011A} & 2011 & 1344 & Recognition & Y & - & Y & - & Outperforms SHOT.  \\
17 & CVFH \cite{Aldoma2012CAD} & 2012 & 308 & Recognition & - & - & - & R & Outperforms VFH. \\
18 & OUR-CVFH \cite{Aldoma2012OUR} & 2012 & 308 & Recognition & - & - & - & - & Outperforms VFH, CVFH, ESF.\\
19 & SH \cite{Behley2012} & 2012 & Vary & Classification & Y & Y & Y & R & Outperforms SHOT in urban environments. \\
20 & COV \cite{Fehr2012Compact} & 2012 & 36 & Recognition & Y & - & - &- &Outperforms SI on real world and synthetic data.\\
21 & SURE \cite{Fiolka2012SURE} & 2012 & 70 & Matching & Y & Y & Y & V, I& Ourperforms NARF descriptor. \\
22 & 3DSSIM \cite{Huang2012} & 2012 & 288 & Matching & Y & Y & - & R & Be robust to various transformations on LiDAR data. \\
23 & GPLF \cite{Hwang2012Robust} & 2012 & 128 & Recognition & - & - & Y &  I & Outperforms SIFT, FPFH in terms of recognition on WOD. \\
24 & GSH \cite{Madry2012Improving} & 2012 & Vary & Classification & Y & -& - & P, V & Outperfors the VFH in PSB and SOC datasets.\\
25 & ESF \cite{Wohlkinger2012Ensemble} & 2012 & 640 & Classification & Y & - & - & - &78.3\% on WOD and can handle data with hole, outliers and coarse object boundaries.\\
26 & RoPS \cite{Guo2013Rotational} & 2013 & 135 & Recognition & Y & Y & Y & R, S &Outperforms SI, SHOT, Thrift. \\
27 & GFH \cite{Chen2014Performance} & 2014 & 864 & Recognition & - & - & - & R & Outperforms SI on SUOD.   \\
28 & Multi-feature \cite{Cheng2014Robust} & 2014 & 59 & Vehicle detection & Y & - & - & 
- &Achieves superior performance in the real urban environment. \\
29 & MCOV \cite{Cirujeda2014MCOV} & 2014 & 6 & Scene analysis & Y & Y & - & R &Outperform the CSHOT when dealing with data with noise or density variation.. \\
30 & LGS \cite{Hadji2014Local} & 2014 & 90 & Classification & Y & - & Y & - &Outperforms SI, FPFH, SHOT on the Stereo dataset. \\
31 & HOPC \cite{Rahmani2014HOPC} & 2014 & - & Action recognition & Y & - & - & V & Shows state-of-the-art performance in MSRAction3D, MSRGesture3D and ActionParis3D datasets. \\
32 & HIGH \cite{Zhao2014Height} & 2014 & 32 & Classification & - & - & - & - &83.7\% on WD, outperforms SI, SHOT, FPFH, 3DSC in terms of class-average accuracy.  \\
33 & B-SHOT \cite{Prakhya2015B} & 2015 & 352 & Matching  & - & - & - & R, T, S &Show comparable performance to SHOT, FPFH, RoPS with lower computational and memory cost in terms of matching.   \\
34 & Point-based \cite{Wang2015A} & 2015 & 18 & Classification & Y & Y & Y & - &Be robust to scale and suitable for classification of objects with small size in TLS point clouds.\\
35 & GOOD \cite{Kasaei2016GOOD} & 2016 & 75 & Recognition & Y & Y & - & P, S &94.0\% on WOD, better than ESF, VFH, GRSD, GFPFH.\\
36 & RISAS \cite{Li2016RISAS} & 2016 & 192 & Matching & - & - & - & V, I, S, R & Outperforms CSHOT on RGB-D scene dataset. \\
37 & GASD \cite{Lima2016} & 2016 & 984 & Recognition & - & - & - & P, S &Outperform ESF, VFH, CVFH, OUR-CVFH in terms of recognition and efficiency on CD.\\
38 & CoSPAIR \cite{Logoglu2016CoSPAIR}& 2016 & 378 & Recognition & - & - & - & - &96.15\% on WOD, 81.46\% on BD in terms of category-level recognition.\\
39 & SGC \cite{Tang2016Signature}& 2016 & 1024 & Matching & Y & Y & Y & - & Outperform SI, 3DSC, SHOT in terms of comprehensive performance.\\
40 & LFSH \cite{Yang2016A}& 2016 & 30 & Registration & Y & Y & - & P & Better than SI, PFH, FPFH, SHOT in terms of dimensionality. \\
\rowcolor{yellow}
41 &  SIPF\cite{Lin2017S} & 2017 & 10 & Detection & Y & Y & - & R, S & Better than SI, PFH, FPFH, SHOT in terms of scale invariant.\\
\rowcolor{yellow}
42 & 3DHoPD \cite{Prakhya20173DHoPD} & 2017 & 18 & Matching & Y & - & Y & R & Better than SHOT, USC, FPFH, 3DSC in terms of computational cost.\\
\hline
\end{tabular}}
\end{table*}

\begin{itemize}
    \item[$\bullet$]\hl{A well-designed hand-crafted descriptor usually require knowledge of expertise domain and careful consideration, which should have capability of describing the noticeable geometric features or discriminative characteristics and be robust to noise, varying dense, occlusion, clutter as well as rotation, scale and translation. }
    \item[$\bullet$] \hl{The great majority of traditional descriptors take advantage of the histogram representation to accumulate geometric attributes or spatial distribution, while others make full use of signature to measure geometry. }
    \item[$\bullet$] \hl{The exploitation of local reference frame or global reference frame makes descriptors much more invariant to rotation than other methods. Therefore, how to construct a unique, unambiguous reference frame with intensive robustness impacts an significant effect on the comprehensive performance of descriptors. }
    \item[$\bullet$]\hl{The global based methods generally estimate a single descriptor vector encoding the whole input 3D point cloud for scenarios, such as 3D object recognition, geometric categorization, and shape retrieval. Therefore, the success of the global descriptor relies on the observation of the entire geometry of the point cloud of the object, which turns out to be a little more complicated. }
    \item[$\bullet$] \hl{The local descriptor construct features resorting to the geometrical information of the local neighborhood of each keypoint obtained from the point cloud using relevant keypoint-extraction algorithms. So the local descriptors are robust to occlusion and clutter} \cite{Guo2013Rotational} \hl{(suitable for applications, e.g. registration, 3D object recognition, and categorization)}, \hl{which the global counterpart is not. However, the local descriptor is commonly sensitive to the changes in the neighborhoods around keypoints} \cite{Hadji2014Local}. \hl{While the hybrid-based descriptor is the sort of descriptor fusing the essential theorem of local and global descriptors or incorporating both kinds of descriptors together to make use of the advantages of local and global features.}
    \item[$\bullet$] \hl{Although traditional hand-crafted 3D point cloud descriptors dominate the development of point cloud processing for a long time, they never achieve the performance of their 2D counterparts and still fail to handle point cloud with noise, occlusions, clutter, and data missing.}
\end{itemize}

\section{3D point cloud descriptors in deep learning age and discussion}
\label{sec:deep}
The unbelievably strong power of deep learning technology brings breakthroughs in obtaining 3D point cloud descriptors. However, the sparse, irregular, and unordered nature of the 3D point cloud challenges the extension of deep learning architectures to 3D point cloud straightforwardly. Therefore, how to represent a 3D point cloud in a format that can be fed into deep learning processing pipeline becomes the crucial issue with which many research works concern. One commonly used approach is pre-processing the raw 3D point cloud data into a structured, fix-sized representation that deep learning architecture can process. Here, we classify these existing approaches into the feature-based, volumetric, multi-view, kd-tree, graph, and point cloud methods according to the representation they adopted.  

\subsection{Feature-based representation}
Feature-based methods first convert the 3D point cloud into a vector by extracting traditional hand-crafted geometric features. \hl{Then, a deep neural network takes these features as input to further process for different applications (e.g. 3D object recognition or 3D reconstruction).} 

In order to learn local geometric features for registration, Khoury et al. \cite{khoury2017learning} resorted to spherical histograms around each point to parameterize the input 3D point cloud by the distribution of points into the bins of subdivision of sphere along the radial, elevation, and azimuth orientation. A deep network is trained to map the high dimensional space of designed histograms into Euclidean space with lower dimensionality by optimizing a triplet loss. Experiments show the effectiveness and high-efficiency of the learned features, which can be used for a drop-in replacement for state-of-the-art hand-crafted descriptors in the existing pipeline of applications.  

\textbf{PPFNet} \cite{deng2018ppfnet} represents the unorganized point cloud as a set of points, normals and point pair features for local geometry or a local patch description. N such local patches are fed into the proposed point pair feature network , where mini-PointNets module for learning local features are integrated into a global feature by a max-pooling layer, and then the concatenation of the global and each local feature is processed by a group of MLPs to produce the global-context aware local descriptor. The whole network is trained by optimizing a newly defined N-tuple loss function to solve the combinatorial problem. Experimental results show that PPFNet obtains outstanding performance in terms of matching accuracy, speed, and robustness to point density and changes in pose. However, it fails to achieve full invariance and has a memory bottleneck. Therefore, Deng et al. \cite{deng2018ppf} proposed \textbf{PPF-FoldNet} based on the idea of PPFNet, PointNet and FoldingNet \cite{yang2018foldingnet}. Unlike PPFNet, they only viewed the PPFs as the local patch here. The PPF-FoldNet adopts an auto-encoder model composing of a PointNet-like encoder with skip links encoding the PPFs as codeword of size 512 and a FoldingNet-like decoder reconstructing the full PPFs, trained in an unsupervised manner. This learned descriptor is 6DoF transformation invariant and powerful.

\textbf{3DmFV-Net}, proposed by Ben-Shabat et al. \cite{Ben20183DmFV}, modifies the fisher vector\cite{sanchez2013image} to formulate the 3D modified fisher vector (3DmFV) for describing point cloud in two ways: exploiting a GMM with Gaussian centered on a 3D grid to preserve the point set structure and adding maximum and minimum function in addition to fisher vector summation to characterize the point sets. Then learned descriptors from 3DmFV by a new CNN are robust to various noise.  

\begin{figure*}[!t]
\centering
\includegraphics[width=\textwidth]{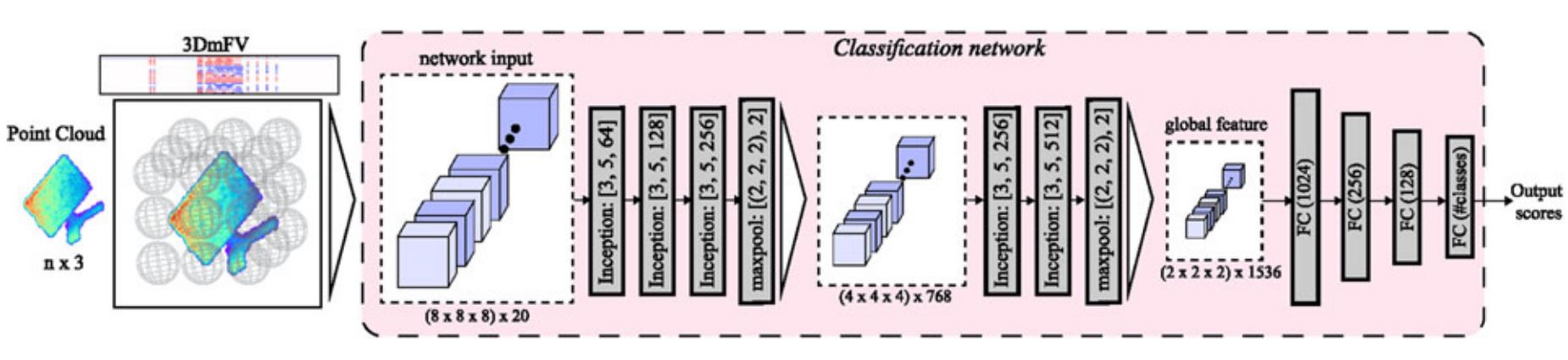}%
\caption{ The architecture of 3DmFV-Net\cite{Ben20183DmFV} .}
\label{fig_3DmFVNet}
\end{figure*}

\subsection{\hl{Voxelization representation}}
Voxelization representation is a straightforward way to voxelize the unstructured 3D point cloud into a regular format of the 3D occupancy grid on which the standard CNNs can process \cite{wu20153d} for applications, such as 3D object recognition and 3D segmentation. There are three commonly used models, namely binary occupancy grid, density grid, and hit grid, which can be adopted \cite{maturana2015VoxNet}. The reason for utilizing this type of methods are as follows. First, the 3D grid is a kind of closest approximation of the 3D world, containing far richer information compared to 2D images. Second, its simple and efficient data structure makes the manipulation and storage easier \cite{wu20153d}. 

\textbf{VoxNet} developed by Maturana et al. \cite{maturana2015VoxNet} is a milestone towards real 3D learning. They applied a 3D convolutional neural network on probabilistic occupancy grids converted from input point cloud for object recognition. Since this representation is not invariant to rotation, they proposed a data augmentation strategy to address this problem. The great success of this approach motivates the development of deep learning on 3D point cloud. However, the democratization into the grid leads to a sparsity problem due to existing of unoccupied space. The sparsity brings a mass of operations of multiplications by zero for dense 3D convolution that is computationally expensive. Engelcke et al. .\cite{engelcke2017vote3deep} designed sparse convolutional layers using a feature-centric voting scheme \cite{wang2015voting} that guarantees filter only applying to occupied grids. The usage of the L1 penalty further increases the sparsity in the intermediate representations. The established \textbf{Vote3Deep} based on them achieves state-of-the-art performance. However, it is challenging to deal with a large point cloud.  

\textbf{VoxelNet}: The region proposal network (RPN)\cite{ren2015faster} for 3D detection task requires dense data while the LiDAR point cloud is sparse, in order to bridge the gap between them, Zhou et al. \cite{zhou2018voxelnet} proposed a VoxelNet architecture. The first block is performing voxel partition on the point cloud, grouping, random sampling operations followed by a newly designed stacked voxel feature encoding that encodes local 3D shape information from each voxel into a list of voxel-wise features. The second block integrates these features into a high-dimensional volumetric representation, which is processed by the third block, the RPN, to generate the descriptor for detection. The whole network is trained in an end-to-end manner and achieves promising results.

\subsection{Multi-view representation}
In the field of computer vision, one fundamental problem is to reason 3D information from 2D images, which motivates the appearance of multi-view learning-based methods. Multi-view representation is one of the simplest ways to represent the 3D point cloud as a collection of 2D views captured from different virtual cameras. The idea behind this kind of approach is firstly mapping the raw 3D point cloud into a set of 2D views and then modeling feature descriptors by drawing from each view individually with 2D CNNs. \hl{Furthermore, the fusion of multi-view is reported to boost the discriminative power of 3D descriptors further for various tasks (e.g. 3D object recognition, 3D object detection and 3D shape retrieval). }   

\begin{figure}[!t]
\centering
\includegraphics[width=4in]{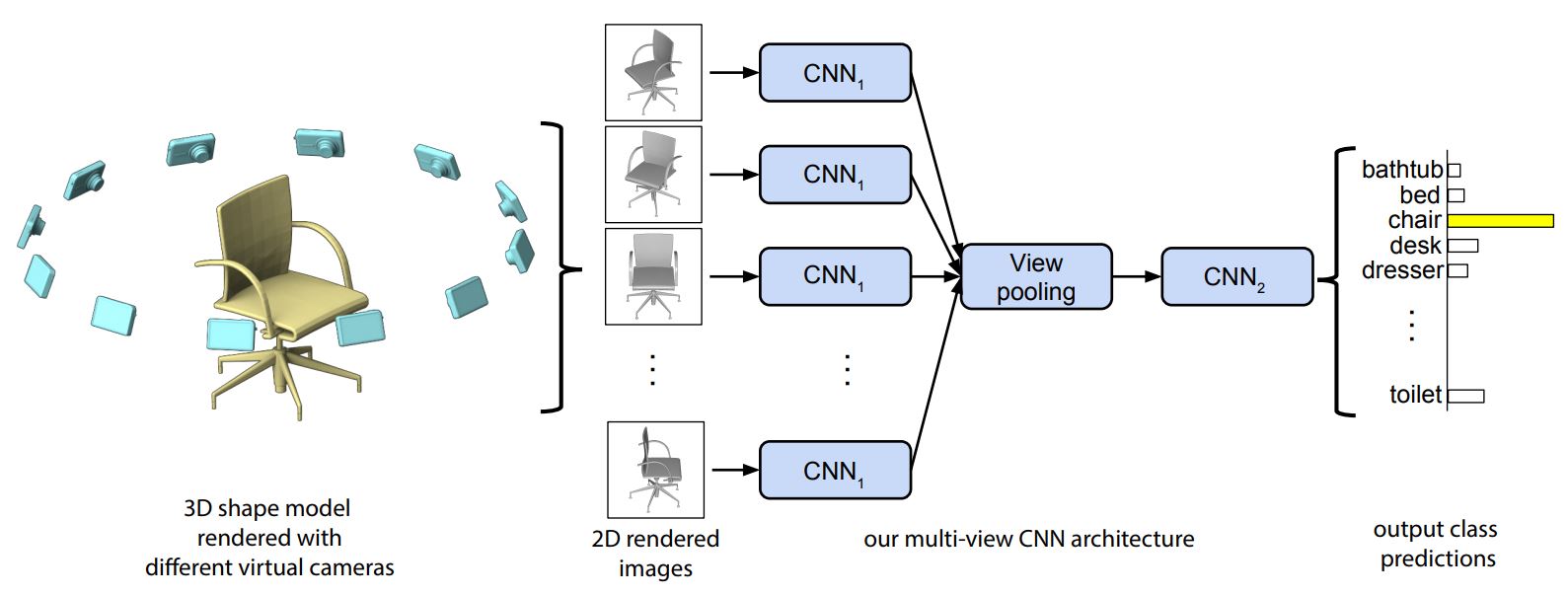}%
\caption{ The architecture of multi-view CNN for 3D object recognition \cite{su2015multi}.}
\label{fig_view}
\end{figure}

Inspired by this idea, \textbf{MVCNN}, as one of the earliest multi-view convolutional neural networks (Figure \ref{fig_view}), is proposed by Su et al. \cite{su2015multi} for recognition and retrieval tasks. The phong reflection model \cite{phong1975illumination} is used to generate the rendering views of the 3D model on each of which the VGG-M network \cite{chatfield2014return} is trained individually for producing informative view-based descriptor. Finally, aggregating the combination of these features from multiple views into a single and compact representation for the 3D object by a novel CNN provides state-of-the-art recognition performance.

\textbf{MV3D} \cite{chen2017multi}, a novel multi-view 3D object detection network, is preset by Chen et al. aiming at 3D object detection in LIDAR point cloud for autonomous driving application. They projected the sparse 3D LIDAR point cloud into a bird's eye view encoded by height, intensity, density, and the front view with three-channel features: height, distance, and intensity. Then the proposed MV3D consisting of a 3D Proposal Network and a Region-based Fusion Network takes these two views together with an image as input. The 3D proposal network produces 3D box proposals from bird's view first, which can be mapped into three views (bird's eye view, front view, and the image plane). Then the region-wise features with the same length via ROI pooling \cite{Girshick2015Fast} from each view are fused to accomplish category classification and 3D box regression tasks jointly. Experimental results on KITTI \cite{geiger2012we} exhibit the remarkable performance of the MV3D network. 

\textbf{Fuseption-ResNet}: Since the view pooling used in MVCNN might smooth out the subtle local patterns, so Zhou et al .\cite{zhou2018learning} put forward a novel view fusion strategy, named Fuseption-ResNet, in which the view-pooling operation serves as a shortcut connection to obtain most robust responses while an Inception-style \cite{szegedy2016rethinking} architecture called Fuseption as residual branch to learn the residual mapping. This framework takes the multi-view feature maps learned from the local path of each view by multiple feature networks with three branches, a variation of MatchNet \cite{han2015matchnet}, as input and outputs a single representation for 3D point cloud registration. Experiments on the public datasets verified its superior performance. 

\textbf{GVCNN}: Another issue arising from MVCNN is that the unexplored content relationship and the discriminative information of the views constrain the performance of descriptors, so Feng et al. \cite{feng2018gvcnn} proposed a \textbf{group-view convolutional neural network}, a hierarchical view-group-shape framework. First, a fully convolutional network is used to extract view level descriptors from each view. Then, a designed grouping module is used to accomplish the estimation of content-based discrimination scores, classification of each view into different groups, and the formulation of group-level descriptors with associated weights. Finally, a weighted fusion process is performed for a combination of all group-level descriptors to form the shape level descriptor.

\textbf{DeepCCFV}: These MVCNN-like methods generally make full use of predefined camera positions, which gives rise to massive over-fitting problems and decreases the generalization to camera constraint-free applications. Therefore, Huang et al. \cite{huang2019deepccfv} designed a DropMax technology that drops the top-k highest stimulates of the output layer at a probability of p to solve the issue. The assembly of the DropMax module into an MVCNN forms a camera constraint-free multi-view convolutional neural network. Furthermore, its applications in single-model and cross-modal retrieval demonstrate the effectiveness and robustness of DeepCCFV. 

\hl{Which views are much more informative? How does the views affect the performance of corresponding tasks? These are two crucial issues into which should be given insight for MVCNN-like approaches. From the discussion above, it can be obviously seen that MVCNN treats all views equally to generate the 3D shape descriptors, which actually limits the performance of these descriptors, while GVCNN takes the content and discriminativity of different views into consideration. That is, similar views should make similar contribution to shape descriptors, 
whereas diverse views have different level impact on the descriptors. In terms of number of views, experimental results show that the overall performance of networks drops sharply as the number of views decreases due to incomplete feature set under this situation, whilst too many views would lead to time consumption. Therefore, the optimal selection of views should make a trade-off between performance and time efficiency and also relies heavily on the requirement of tasks. Generally, 12 views sampled with 30 degrees interval around a horizontal circle are one of frequent choices for MVCNN-like methods.
}

\subsection{Kd-tree representation}
The approaches described above either cause much higher memory and time cost or capture insufficient 3D nature of point cloud for various tasks. Kd-tree as 3D point cloud representation is proposed to solve these problems as well as unstructured input issues. This kind of method first makes full use of a kd-tree indexing structure (integrating post-processing on the kd-tree structure) to create a hierarchical and locality-preserving order of the points. \hl{Then, a designed deep architecture deals with representation based on this structure to extract descriptors for corresponding applications. }

\textbf{Deep Kd-Network} proposed by Klokov et al.\cite{klokov2017escape} is the pioneering work on this field. Their method utilizes kd-tree structure to employ  parameters sharing and compute the hierarchical representations from leaves to root by a feed-forward network. Although the Kd-Network achieves competitive performance, it is not robust to the rotation. The \textbf{3DContextNet} introduced by Zeng et al. \cite{zeng20183dcontextnet} consists of feature learning and aggregation stages progressively along with the kd-tree structure. In the feature learning stage, local and global contextual cues are learned to encode point features followed by the aggregation step estimating the representation vector hierarchically in a bottom-up manner. This method is competitive. Gadelha et al. \cite{gadelha2018multiresolution} represented the kd-tree-structured point cloud as a group of 1D ordered list of points at multi-resolution levels, which is fed into a flexible auto-encoder architecture (named multiresolution tree network, \textbf{MRTNet}) to learn coarse-to-fine descriptors for performing different tasks, e.g. 3D ojbect classification.

\begin{figure}[!t]
\centering
\includegraphics[width=4in]{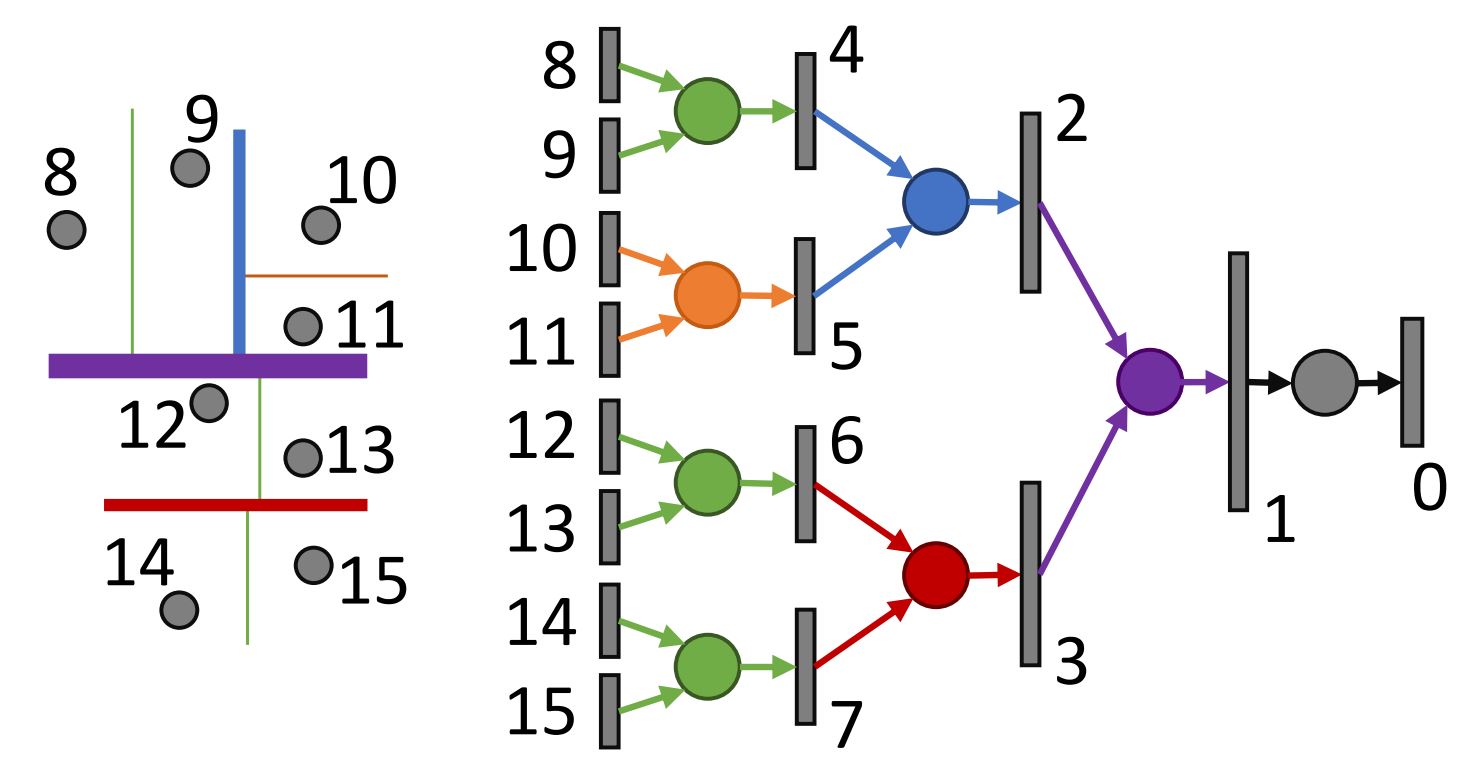}%
\caption{ A kd-tree structure and associated Kd-network \cite{klokov2017escape}.}
\label{fig_kdtree}
\end{figure}

\subsection{Point cloud input }
Point cloud provides an intuitive, flexible, memory efficiency and expressive representation for 3D geometry, which actually can make the deep learning architecture learn underlying features readily. On the other hand, the prior types of methods on conversion into regular representation encounter information loss and quantitation artifacts leading to unsatisfactory performance. Recent works, therefore, shift attention toward frameworks operating on raw 3D point cloud directly instead of intermediate regular representation. 

\begin{figure*}[!t]  
\centering
\includegraphics[width=\textwidth]{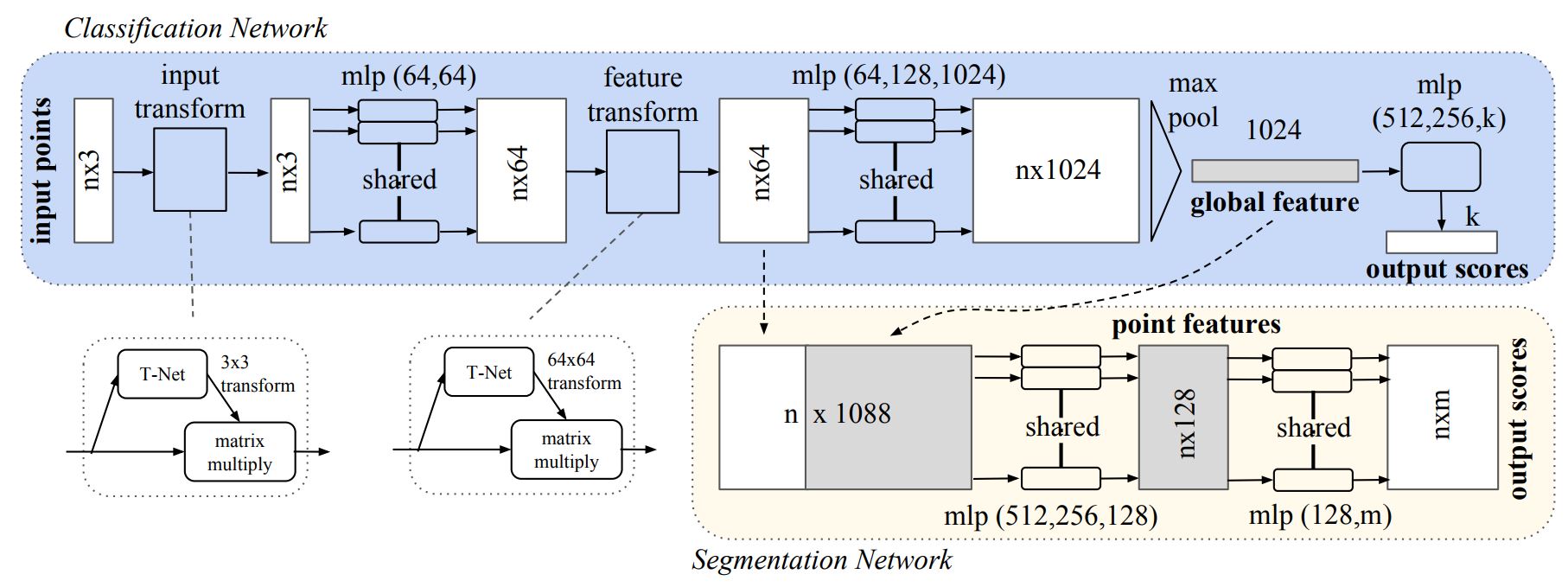}%
\caption{ The architecture of PointNet \cite{Charles2017PointNet}.}
\label{fig_pointnet}
\end{figure*}

\textbf{PointNet} \cite{qi2017pointnet} (Figure \ref{fig_pointnet}) is the first work using a deep learning to deal with irregular point cloud straightforwardly. The idea behind this network is to aggregate all the learned point-wise features into a global representation signature by max pooling operation which is served as the symmetric function to make the model invariant to the input permutation. In addition, the learned representation should be not sensitive to geometric transformations of point cloud, consequently PointNet adopts a T-net to impose an affine transformation on input points. Experiments show that the PointNet can achieve state-of-the-art performance in 3D classification and segmentation tasks. However, it fails to sufficiently capture the local structure in 3D space \cite{qi2017pointnet++} due to lack of ConvNet-like hierarchical feature aggregation \cite{li2018so}. In order to resolve this issue, an modified version of PointNet with hierarchical structure, termed \textbf{PointNet++} \cite{qi2017pointnet++} is introduced. This network can extract features from different scales hierarchically by progressively repeating the following operations across different sizes of local regions: selecting points using iterative farthest point sampling in sampling layer as centroids for local regions, determining the neighborhoods of centroid points as the local region and using a mini-PointNet to extract features from these regions. Although these contributions enable the PointNet++ to produce remarkable results on point cloud benchmarks, the recursive operations slow the inference speed and it is still treat the points independently \cite{wang2018dynamic}. And both PointNet and PointNet++ fail to take the spatial relationship amongst individual points explicitly. 

 \textbf{PointCNN} \cite{li2018pointcnn}, unlike PointNet, builds a $\chi -Conv$ block to solve unordered problem. This block first applies MLP to learn a $K\times K$ $\chi-$ transformation from K selected points, $\chi = MLP\left (p_{1},p_{2},...,p_{k}  \right )$, followed by weighting and permuting operations simultaneously. Then, performing convolution on the transformed features. Analogous to deep grid-based CNNs, PointCNN uses the $\chi-Conv$ block instead of Conv layer to construct its architectures. Experiments demonstrate excellent performance on multiple benchmarks and tasks. 

\textbf{RSNet}: For 3D segmentation task, PointNet does not consider local information and PointNet++ costs heavily extra computations. To handle these problems, Huang et al. \cite{huang2018recurrent} developed a local dependency module to model local context, which includes high-efficiency Slice Pooling Layer converting the features of unordered input points into ordered sequence of features by grouping and aggregating operations, RNN Layer modeling dependencies in the sequence and Slice Unpooling Layer reversely projecting the updated features to each point with time complexity of O(1) w.r.t the local context resolution. This local dependency module together with an input feature extraction module constitute the \textbf{RSNet} which surpasses the PointNet and PointNet++ on S3DIS \cite{armeni20163d}, ScanNet \cite{dai2017scannet} and ShapeNet \cite{yi2016scalable}. 

Since the representation of output is also an important consideration for segmentation, as a result, Wang et al. \cite{wang2018sgpn} introduced \textbf{Similarity Group Proposal Network (SGPN)} employing PointNet/PointNet++ to learn expressive feature for each point which is input into three individual branches to produce a similarity matrix, confidence map and semantic prediction. The similarity matrix that indicates similarity between point pairs by measuring distance is combined with the confidence scores to yield group proposal for each point. However, this method can be applied to large point cloud due to the fact that increasing the number of points will scale the size of the similarity matrix. Similarly, Associatively Segmenting Instances and Semantics proposed by Wang et al. \cite{wang2019associatively} is also a multi-task architecture consisting of a baseline auto-encoder model followed by a new ASIS module to perform semantic-aware instance segmentation and instance-fused semantic segmentation simultaneously. These two tasks benefit from each other to improve performance.  

\subsection{Graph-based representation}
This type of approaches usually construct a representation graph G = (V,E) for either each point and its neighbors determined by k-nearest neighbor (k-NN) strategy to represent the local structure of point cloud or the entire point cloud to reflect relationship among points, where V and  $E\subseteq V\times V$ are vertices and edges. Then, the deep convolution network is constructed to learn descriptive relationship features among points.    

\textbf{GCNN}: From the previous discussion, it is stated that PointNet and PointNet++ fail to capture the geometric relationships among points. As a result, in order to fully exploit local geometric structure while maintaining permutation invariance, Wang et al. \cite{wang2018dynamic} defined EdgeConv operators, the key ingredient of \textbf{Dynamic Graph CNN}, by imposing channel-wise symmetric aggregation operation on edge features formulated for k-NN graph constructed for each point and its neighbors. They assembled the EdgeConv into PointNet to perform classification and segmentation tasks. While Wang et al. \cite{wang2018local} designed a spectral graph convolution (contains Graph Fourier transform, Spectral modulation, Feature filtering and Reverse Fourier transform steps) for PointNet++ framework processing the k-NN graph for each point to capture local underlying structure. Meanwhile, they substituted the novel graph pooling strategy for standard pooling operation to aggregate features from locally spectral-space graph in a recursive manner.  

Different from aforementioned approaches, Landrieu et al. \cite{landrieu2018large} proposed a superpoint graph structure to capture the relationship among instance parts in 3D point cloud scene to resolve the problem of limitation of size of input. They first used a global energy model \cite{guinard2017weakly} to geometrically split the point cloud into semantically homogeneous elements, referred to as superpoints, which allows to construct a superpoint graph by connecting adjacent superpoints with superedges. Finally, PointNet is employed to embed each sueprpoint into a vector with fixd-size which is fed into a deep learning algorithm with graph convolution to perform segmentation in an end-to-end manner. 

\subsection{Multi-sensors}
In the scenario using both cameras and 3D sensors, multi-sensors based approaches arise to make full use of advantages of different modalities. And these methods usually utilize a complicated pipeline to process 2D images and 3D point cloud sequentially or parallelly for carrying out reliable point cloud analysis in a joint deep learning network architecture.

\textbf{3DMV} is proposed by Dai et al. \cite{dai20183dmv} to address the integration problem of multiple modalities, which takes the 3D representation as well as 2D image as input to infer 3D semantics in a joint, end-to-end manner. In this framework, they first used a 2D CNN to extract 2D feature maps from multiple RGB images. Then these features were mapped into 3D space through a differentiable backprojection layer. Finally, both the backprojected features and 3D input were fed into a 3D CNN to perform semantic segmentation. Joining the benefits of existing approaches actually brings significant performance. However, the usage of voxel grid leads to high computation cost. Different from 3DMV, Jaritz et al. \cite{jaritz2019multi} chose to work in point cloud space directly to exploit the fusion of multiview and geometry information. And this \textbf{MVPNet} achieved superior performance against the 3DMV.

\subsection{\hl{Summary of Deep Learning based Descriptor}}
Table \ref{tab:Deeplearningforpoincloud} presents the properties of some representative deep learning-based approaches. Table \ref{tab:recognition}, table \ref{tab:PartSegmentation}, table \ref{tab:SemanticSegmentation} shows the performance of several state-of-the-art. 

\begin{itemize}
    \item[$\bullet$] \hl{From the previous discussion, it can be clearly found that the feature-based methods are mainly used in applications, such as 3D object recognition and 3D reconstruction. And yet this kind of approaches require to extract the traditional hand-crafted descriptor initially, which inevitably constrains the overall performance of these methods to some extend due to the inherent drawbacks and limited descriptive power of the descriptors.}
    \item[$\bullet$] \hl{Voxelization representation is a representation that can bridge the gap between 2D and 3D worlds. And its compatibility with 3D convolution networks (3D ConvNets) makes it an intuitively straightforward choice to convert 3D point cloud into voxelization }\cite{li2018so}. \hl{However, (1) the nature of 3D ConvNets makes these methods much more computationally expensive. (2) And since voxel grid representation needs to represent both free, occupied, and unknown space, therefore it usually consumes much higher memory usage, especially in rare cases} \cite{qi2016volumetric}. \hl{(3) On the other hand, the appearance of sparsity, excessive storage consuming, and expensive computation cost further impose a limitation on the resolution of voxelization representation, causing  artifacts} \cite{wang2018dynamic}\cite{qi2017pointnet}, \hl{which makes these methods not well suited for scenarios requires capturing fine details or high-resolution.}
    \item[$\bullet$] \hl{Multi-view methods turn the 3D problem into a 2D problem, which allows the well-established 2D CNNs developed for image processing to be leveraged directly. Therefore, compared to voxelization representation, multi-view representation improves computational efficiency significantly. Nevertheless, a fixed number of 2D images is not an ideal approximation of 3D shape due to the fact that 2D projection suffers from surface information loss}  \cite{su2018splatnet}. \hl{And it is difficult to determine the appropriate number of views selection for modeling underlying 3D structure. Since the insufficient number of views may not reflect the character of the 3D point cloud, while too many images will cause unwanted time-consuming.} 
    \item[$\bullet$] \hl{In reference to Kd-tree based representation, 3D object recognition or classification and segmentation are common application scenarios. Without transforming 3D point cloud into other regular representations, this kind of methods can capture the latent relations between areas and fully exploit the nature of 3D itself by the introduction of kd-tree structure, which reduces the memory footprint and computational cost. However, since the kd-tree structure is sensitive to rotations, so most of these descriptors are not invariant to rotation transformation.} 
    \item[$\bullet$] \hl{Point-based methods are currently becoming mainstream research direction due to its superior characteristics, which achieve state-of-the-art results for deep learning based tasks (e.g. 3D object recognition, 3D segmentation and 3D object detection). However, almost all of these methods can only deal with small scale point cloud with fixed number of points or fixed size (e.g. 1024 or 2048 points), which limits the extension to applications involving large scale point cloud. }
    \item[$\bullet$] \hl{Compared to Voxelization or Multi-view representations, graph-based representation is more suitable for irregular data processing scenarios (e.g. 3D object recognition and segmentation) and is capable of taking advantage of the sparse nature of point cloud. However, how to design effective convolution, deconvolution, pooling, unpooling modules for graph-based CNN is still a challenging study and worth further investigation.}
    \item[$\bullet$] \hl{As for multi-sensors based approaches, 2D image provides much richer texture information better for discriminating objects with physically similar appearance but different texture, while 3D point cloud offers shape information suitable for semantically similar things with different scales} \cite{Munoz2012Co}. \hl{The fusion of these two kinds of data, therefore, is beneficial to the improvement of the representative power of the learned features for segmentation applications. However, fusion data from 2D and 3D domains still remains a challenging problem due to the fact that they live in different data spaces and there is no exact one-to-one correspondence between them. }
    
\end{itemize}

\begin{table*}[]
\caption{Properties of deep learning methods on point cloud processing.}
\label{tab:Deeplearningforpoincloud}
\centering
\resizebox{\textwidth}{!}{
\begin{tabular}{|c|c|c|>{\columncolor{yellow}}c|c|c|>{\columncolor{yellow}}c|>{\columncolor{yellow}}c|>{\columncolor{yellow}}c|c|c|}
\hline
No.                 & Method                                   & Year                  & Application        & Category                 & Type                          & Parameters(million) & Train Time(hour) & Test Time(ms) & GPU                          & Implementation                \\
\hline
1                   & 3DShapeNet \cite{wu20153d}    & 2015                  & Object Recognition & Voxelization               & 3D CNN                        & -                   & -                & -             & Tesla K40                    & C++, Matlab                 \\
\hline
2                   & VoxNet \cite{maturana2015VoxNet}         & 2015                  & Object Recognition & Voxelization               & 3D CNN                        & 0.9                 & 6-12             & 6             & Tesla K40                    & C++, Python                 \\
\hline
3                   & SEGCloud \cite{tchapmi2017segcloud}      & 2017                  & Segmentation       & Voxelization               & 3D-FCNN                       & -                   & -                & -             & -                            & Caffe                       \\
\hline
4                   & VoxelNet \cite{zhou2018voxelnet}         & 2018                  & Object Detection   & Voxelization               & 3D CNN                        & -                   & -                & 225           & Titan X                      & Tensorflow                  \\
\hline
\multirow{2}{*}{5}  & \multirow{2}{*}{OctNet \cite{riegler2017octnet} }                 & \multirow{2}{*}{2017}  & Segmentation       & \multirow{2}{*}{Octree}  & CNN                           & -                   & -                & -             & \multirow{2}{*}{-}                           & \multirow{2}{*}{Torch}        \\ \cline{4-4} \cline{6-9}
                    &                                          &                       & Object Recognition &                          & U-Net                         & -                   & -                & -             &                             &                          \\
\hline
\multirow{2}{*}{6}  & \multirow{2}{*}{O-CNN \cite{wang2017cnn} }               & \multirow{2}{*}{2017} & Segmentation       & \multirow{2}{*}{Octree}  & CNN                           & -                   & -                & -             & \multirow{2}{*}{Gefore 1080} & \multirow{2}{*}{Caffe}      \\ \cline{4-4} \cline{6-9}
                    &                                          &                       & Object Recognition &                          & CNN+CRF                       & -                   & -                & 39.5          &                              &                              \\
                    \hline
\multirow{2}{*}{7}  & \multirow{2}{*}{$\Psi$-CNN \cite{lei2019octree}} & \multirow{2}{*}{2019} & Segmentation       & \multirow{2}{*}{Octree}  & \multirow{2}{*}{3D CNN}       & -                   & -                &               & \multirow{2}{*}{Titan XP}    & \multirow{2}{*}{Tensorflow}  \\ \cline{4-4} \cline{7-9}
                    &                                          &                       & Object Recognition &                          &                               & -                   & -                & 34.1          &                              &                              \\ 
                    \hline
\multirow{2}{*}{8}  & \multirow{2}{*}{Kd-Net \cite{klokov2017escape} }    & \multirow{2}{*}{2017} & Segmentation       & \multirow{2}{*}{Kd-tree} & CNN                           & -                   & -                & -             & \multirow{2}{*}{Titan X}     & \multirow{2}{*}{Theano}      \\ \cline{4-4} \cline{6-9}
                    &                                          &                       & Object Recognition &                          & Encoder-decoder               & -                   & 16               & -             &                              &                              \\
                    \hline
9                  & PCCN \cite{wang2018deep}                                   & 2018                  & Segmentation       & Point                    & DNN                           & 74.67               & -                & 61            & GTX 1080Ti                   & -                            \\
\hline
10                  & MVCNN \cite{su2015multi}                                   & 2015                  & Object Recognition & Multi-view               & Multi-branch                  & 60                  & -                & -             & -                            & Pytorch                     \\
\hline
11                  & MV3D \cite{chen2017multi}                     & 2017                  & Object Detection   & Multi-view               & Multi-branch                  & 103                 & -                & 0.36          & Titan X                      & Tensorflow                  \\
\hline
12                  & GVCNN \cite{feng2018gvcnn}                                 & 2018                  & Object Recognition & Multi-view               & Multi-branch                  & -                   & -                & -             & -                            & -                           \\
\hline
13                  & PPFNet \cite{deng2018ppfnet}                                  & 2018                  & Registration       & Feature                  & Multi-branch                  & -                   & -                & 2.25          & Titan X                      & Tensorflow                  \\
\hline
14                  & FoldingNet \cite{yang2018foldingnet}                             & 2018                  & Reconstruction     & Feature                  & Encoder-decoder               & -                   & -                & -             & -                            & -                           \\
\hline
15                  & 3DmFV-Net \cite{Ben20183DmFV}       & 2018                  & Object Recognition & Feature                  & DNN                           & 4.6                 & 7                & 3.7           & Titan XP                     & Tensorflow                  \\
\hline
16                  & PPF-FoldNet \cite{deng2018ppf}           & 2018                  & Reconstruction     & Feature                  & Encoder-decoder               & -                   & -                & 3969          & Titan X                      & Tensorflow                  \\
\hline
\multirow{2}{*}{17} & \multirow{2}{*}{ShapeContextNet \cite{xie2018attentional}}          & \multirow{2}{*}{2018} & Segmentation       & \multirow{2}{*}{Feature} & \multirow{2}{*}{CNN}          & -                   & -                & -             & \multirow{2}{*}{-}           & \multirow{2}{*}{-}          \\ \cline{4-4} \cline{7-9}
                    &                                          &                       & Object Recognition &                          &                               & -                   &                  & -             &                              &                             \\
                    \hline
18 & DeepNet \cite{ravanbakhsh2016deep} & 2017 & Object Recognition & Points & - & - & - & - & -& Tensorflow\\
 \hline
\multirow{2}{*}{19} & \multirow{2}{*}{PointNet \cite{qi2017pointnet} }               & \multirow{2}{*}{2017} & Object Recognition & \multirow{2}{*}{Point}   & \multirow{2}{*}{MLP-based}    & 0.8                 & 3-6              & 11.6          & \multirow{2}{*}{GTX 1080}                     & \multirow{2}{*}{Tensorflow}                  \\ \cline{4-4}\cline{7-9}
                    &                                          &                       & Segmentation       &                          &                               & 3.5                 & 6-12             & 25.3          &                              &                             \\
                    \hline
\multirow{2}{*}{20} & \multirow{2}{*}{PointNet++ \cite{qi2017pointnet++}}              & \multirow{2}{*}{2017} & Segmentation       & \multirow{2}{*}{Point}   & MLP-based                     & -                   & -                & -             & \multirow{2}{*}{GTX 1080}    & \multirow{2}{*}{Tensorflow} \\ \cline{4-4} \cline{6-9}
                    &                                          &                       & Object Recognition &                          & Encoder-decoder               & 8.7                 & 20               & 82.4          &                              &                             \\
                    \hline
\multirow{2}{*}{21} & \multirow{2}{*}{PCNN \cite{atzmon2018point}}                    & \multirow{2}{*}{2018} & Segmentation       & \multirow{2}{*}{Point}   & Encoder-decoder               & 5.4                 & 52.4             & 200           & \multirow{2}{*}{P100}        & \multirow{2}{*}{Tensorflow}  \\ \cline{4-4} \cline{6-9}
                    &                                          &                       & Object Recognition &                          & CNN                           & 8.1                 & 25               & 80            &                              &                             \\
                    \hline
22                 & SSCN \cite{graham20183d}            & 2018                  & Segmentation       & Point                    & FCN or U-Net                  & -                   & -                & -             & -                            & -                           \\
\hline

\multirow{2}{*}{23} & \multirow{2}{*}{PointCNN \cite{li2018pointcnn}}                & \multirow{2}{*}{2018} & Segmentation       & \multirow{2}{*}{Point}   & \multirow{2}{*}{CNN}          & -                   & -                & -             & \multirow{2}{*}{Tesla P100}  & \multirow{2}{*}{Tensorflow}  \\ \cline{4-4} \cline{7-9}
                    &                                          &                       & Object Recognition &                          &                               & 0.6                 & -                & 12            &                              &                              \\
                    \hline
\multirow{2}{*}{24} & \multirow{2}{*}{SO-Net \cite{li2018so}}                  & \multirow{2}{*}{2018} & Segmentation       & \multirow{2}{*}{Point}   & Encoder-decoder               & -                   & -                & -             & \multirow{2}{*}{GTX 1080Ti}  & \multirow{2}{*}{Pytorch}    \\ \cline{4-4} \cline{6-9}
                    &                                          &                       & Object Recognition &                          & ConvNet                       & 11.5                & 3                & 59.6          &                              &                             \\
                    \hline
25                  & SPLATNet \cite{su2018splatnet}        & 2018                  & Segmentation       & Point                    & CNN                           & 1.41                & -                & 9400          & GTX 1080Ti                   & caffe             \\
\hline
26                  & SGPN \cite{wang2018sgpn}           & 2018                  & Segmentation       & Point                    & Multi-task                    & -                   & 16-17            & 170           & GTX 1080Ti                   & Tensorflow                  \\
\hline

\multirow{2}{*}{27} & \multirow{2}{*}{SpiderCNN \cite{xu2018spidercnn}}               & \multirow{2}{*}{2018} & Segmentation       & \multirow{2}{*}{Point}   & \multirow{2}{*}{CNN}          & -                   & -                & -             & \multirow{2}{*}{GTX 1080Ti}  & \multirow{2}{*}{Tensorflow}  \\ \cline{4-4} \cline{7-9}
                    &                                          &                       & Object Recognition &                          &                               & -                   & -                & 71.68         &                              &                             \\
                    \hline
\multirow{2}{*}{28} & \multirow{2}{*}{SRN \cite{duan2019structural}}                     & \multirow{2}{*}{2019} & Segmentation       & \multirow{2}{*}{Point}   & \multirow{2}{*}{MLP-based}    & -                   & -                & -             & \multirow{2}{*}{-}           & \multirow{2}{*}{Tensorflow}  \\ \cline{4-4} \cline{7-9}
                    &                                          &                       & Object Recognition &                          &                               & -                   & -                & -             &                              &                             \\
                    \hline
\multirow{2}{*}{29} & \multirow{2}{*}{RS-CNN \cite{liu2019relation}}                  & \multirow{2}{*}{2019} & Segmentation       & \multirow{2}{*}{Point}   & \multirow{2}{*}{CNN}          & -                   & -                & -             & \multirow{2}{*}{-}           & \multirow{2}{*}{Pytorch}    \\ \cline{4-4} \cline{7-9}
                    &                                          &                       & Object Recognition &                          &                               & 1.41                & -                & -             &                              &                             \\
                    \hline 
 \multirow{2}{*}{30} & \multirow{2}{*}{ Point2Sequence \cite{liu2019point2sequence}}                  & \multirow{2}{*}{2019} & Segmentation       & \multirow{2}{*}{Point}   & \multirow{2}{*}{Encoder-decoder}          & -                   & -                & -             & \multirow{2}{*}{-}           & \multirow{2}{*}{-}    \\ \cline{4-4} \cline{7-9}
                    &                                          &                       & Object Recognition &                          &                               & -                & -                & -             &                              &                             \\
                    \hline                   
\multirow{2}{*}{31} & \multirow{2}{*}{SFCNN \cite{rao2019spherical}}                   & \multirow{2}{*}{2019} & Segmentation       & \multirow{2}{*}{Point}   & \multirow{2}{*}{CNN}          & -                   & $\textless$ 24    & -             & \multirow{2}{*}{GTX 1080Ti}  & \multirow{2}{*}{-}          \\ \cline{4-4} \cline{7-9}
                    &                                          &                       & Object Recognition &                          &                               & -                   & -                & -             &                              &                             \\
                    \hline
32                  & SyncSpecCNN \cite{Li2017SyncSpecCNN}                            & 2017                  & Segmentation       & Graph                    & GCNN                          & -                   & -                & -             & -                            & -                           \\
\hline
33                  & ECC \cite{simonovsky2017dynamic}          &           2017            &       Object Recognition             & Graph                    &   GCNN                            &         -            &          -        &          -     &                 -             &                -             \\
\hline
34                  & SPGraph \cite{landrieu2018large}                                 & 2018                  & Segmentation       & Graph                    & GCNN                          & -                   & -                & 599,000       & GTX 1080Ti                   & Pytorch                     \\
\hline
\multirow{2}{*}{35} & \multirow{2}{*}{KC-Net \cite{shen2018mining}}                  & \multirow{2}{*}{2018} & Segmentation       & \multirow{2}{*}{Graph}   & \multirow{2}{*}{GCNN}         & -                   & -                & -             & \multirow{2}{*}{GTX 1080}    & \multirow{2}{*}{Caffe}      \\ \cline{4-4} \cline{7-9}
                    &                                          &                       & Object Recognition &                          &                               & 0.9                 & -                & 12            &                              &                              \\
                    \hline
\multirow{2}{*}{36} & \multirow{2}{*}{RGCNN \cite{te2018rgcnn}}                   & 2018                  & Segmentation       & \multirow{2}{*}{Graph}   & \multirow{2}{*}{GCNN}         & -                   & -                & -             & \multirow{2}{*}{GTX 1080Ti}  & \multirow{2}{*}{Tensorflow} \\ \cline{4-4} \cline{7-9}
                    &                                          &                       & Object Recognition &                          &                               & 22.4                & -                & 7.5           &                              &                             \\
                    \hline

\multirow{2}{*}{37} & \multirow{2}{*}{Spec-GCN \cite{wang2018local}}                & \multirow{2}{*}{2018} & Segmentation       & \multirow{2}{*}{Graph}   & \multirow{2}{*}{GCNN}         & -                   & -                & 8             & \multirow{2}{*}{}            & \multirow{2}{*}{Tensorflow} \\ \cline{4-4} \cline{7-9}
                    &                                          &                       & Object Recognition &                          &                               & -                   & -                & 12            &                              &                             \\
                    \hline

\multirow{2}{*}{38} & \multirow{2}{*}{DGCNN \cite{wang2018dynamic}}                   & \multirow{2}{*}{2019} & Segmentation       & \multirow{2}{*}{Graph}   & \multirow{2}{*}{GCNN}         & -                   & -                & -             & \multirow{2}{*}{Titan X}     & \multirow{2}{*}{Tensorflow}  \\ \cline{4-4} \cline{7-9}
                    &                                          &                       & Object Recognition &                          &                               & 21                  & -                & 27.2          &                              &                             \\
                    \hline

\multirow{2}{*}{39} & \multirow{2}{*}{Point2Node \cite{han2019point2node}}              & \multirow{2}{*}{2019} & Segmentation       & \multirow{2}{*}{Graph}   & \multirow{2}{*}{GCNN}         & 11.14               & -                & -             & \multirow{2}{*}{-}           & \multirow{2}{*}{-}          \\ \cline{4-4} \cline{7-9}
                    &                                          &                       & Object Recognition &                          &                               & -                   & -                & -             &                              &                             \\
                    \hline
 40 & HPEINet \cite{Jiang2019Hierarchical} & 2019 & Segmentation & Graph & Multi-branch+Encoder-decoder & - & - & - & - & -\\
 \hline
41                  & GACNet \cite{wang2019graph}                                   & 2019                  & Segmentation       & Graph                    & GCNN+Encoder-decoder          & -                   & -                & -             & -                            & -                           \\
\hline

\multirow{2}{*}{42} & \multirow{2}{*}{PointConv \cite{wu2019pointconv}}               & \multirow{2}{*}{2019} & Segmentation       & \multirow{2}{*}{Graph}   & \multirow{2}{*}{Multi-branch} & -                   & -                & 500           & \multirow{2}{*}{GTX 1080Ti}  & \multirow{2}{*}{Tensorflow}  \\ \cline{4-4} \cline{7-9}
                    &                                          &                       & Object Recognition &                          &                               & -                   & -                & -             &                              &                              \\

\hline
\multirow{2}{*}{43} & \multirow{2}{*}{PointWeb \cite{zhao2019pointweb}}                & \multirow{2}{*}{2019} & Segmentation       & \multirow{2}{*}{Graph}   & \multirow{2}{*}{GCNN}          & -                   & -                & -             & \multirow{2}{*}{-}           & \multirow{2}{*}{Pytorch}    \\ \cline{4-4} \cline{7-9}
                    &                                          &                       & Object Recognition &                          &                               & -                   & -                & -             &                              &                             \\ \hline

44                  & 3DMV \cite{dai20183dmv}                                    & 2018                  & Segmentation       & Multi-sensors            & Multi-branch+2D CNN+3D CNN    & 10.2                & 24               & -             & -                            & Pytorch                     \\
\hline
45                  & MVPNet \cite{jaritz2019multi}                                   & 2019                  & Segmentation       & Multi-sensors            & Encoder-decoder               & 0.98                & 12               & 2220          & GTX 1080Ti                   & -                           \\
\hline
\end{tabular}}
\end{table*}

\begin{table*}[!t]
\caption{Comparisons of recognition accuracy(\%) on ModelNet10 and ModelNet40. }
\label{tab:recognition}
\centering
\begin{tabular}{ccccc} \hline
Method & Representation  & Input Size  & ModelNet10  & ModelNet40 \\ \hline
3DShapeNets \cite{wu20153d} & Voxelization  & $30^{3}$ & 83.54\% & 77.32\% \\
VoxNet \cite{maturana2015VoxNet} &  Volumetric & $32^{3}$ & 92.0\%  & 83.0\% \\
OctNet \cite{riegler2017octnet} & Volumetric & $128^{3}$ & 90.9\% & 86.5\% \\
O-CNN \cite{wang2017cnn} & Points+normals & - & - & 90.6\%\\
$\Psi$-CNN \cite{lei2019octree} & Points & - & 94.6\% & 92.0\%\\ 
Kd-Net \cite{klokov2017escape} & Points  & $2^{15} \times 3$ & 93.5\% & 88.5\% \\
MVCNN \cite{su2015multi} & Multi-view & $12\times224^{2}$ & - & 90.1\%  \\
GVCNN \cite{feng2018gvcnn} & Multi-view & 8 $\times$ & - & 93.1\% \\
FoldingNet \cite{yang2018foldingnet} & Points  & $2048\times3$ & 94.4\% & 88.4\%  \\
3DmFV-Net \cite{Ben20183DmFV} & Points & 2048 $\times$ 3 & 95.2 & 91.4\% \\
ShapeContextNet \cite{xie2018attentional} & Points & $1024 \times 3$ & - & 90.0\%\\
DeepNet \cite{ravanbakhsh2016deep} & Points & $5000\times3$ & - & 90.0\% \\
PointNet \cite{qi2017pointnet} & Points & $1024 \times 3$ & - & 89.2\%  \\
PointNet++ \cite{qi2017pointnet++}& Points+normals & $5000\times6$ & - & 91.9\%   \\
PCNN \cite{atzmon2018point} & Points & $1024 \times 3$ & 94.9\% & 92.3\% \\ 
PointCNN \cite{li2018pointcnn} & Points & $1024 \times 3$ & - & 92.5\% \\
SO-Net \cite{li2018so} & Points+normals & $5000\times3$ & 95.7\% & 93.4\% \\
SpiderCNN \cite{xu2018spidercnn} & Points+normals & $1024 \times 3$ & - & 92.4\% \\
SRN \cite{duan2019structural} & Points & $1024\times3$ & - & 91.5\% \\
RS-CNN \cite{liu2019relation} & Points & $1024\times3$ & - & 93.6\% \\
Point2Sequence \cite{liu2019point2sequence} & Points & $1024\times3$ & 95.3\% & 92.6\%   \\
SFCNN \cite{rao2019spherical} & Points+normals & $1024\times6$ & - & 92.3\% \\
ECC \cite{simonovsky2017dynamic} & Points & 1000 $\times$3 &  90.0\%  & 83.2\% \\
KC-Net \cite{shen2018mining} & Points & $1024\times3$  & 94.4\% & 91.0\% \\
RGCNN \cite{te2018rgcnn} & Points & $1024 \times $3& - & 90.5\%\\
Spec-GCN \cite{wang2018local} & Points & $2048 \times 3$ & - & 92.1\%\\
DGCNN \cite{wang2018dynamic}  & Points & $1024 \times 3$ & - & 92.2\% \\
Point2Node \cite{han2019point2node} & Points & $1000 \times 3$ & - & 93.0\%\\
PointConv \cite{wu2019pointconv} & Points & $1024 \times 3$ &- & 92.5\%\\
PointWeb \cite{zhao2019pointweb} & Points+normals & $1024 \times 3$ & - & 92.3\% \\
\hline
\end{tabular}
\end{table*}

\begin{table*}[!t]
\caption{Experimental comparison of part segmentation with the state-of-the-art approaches on ShapeNet part dataset. The mean IoU across all the shape instances and IoU for each category are reported. 'EP' stands for earphone }
\label{tab:PartSegmentation}
\centering
\resizebox{\textwidth}{!}{
\begin{tabular}{c|c|cccccccccccccccc}
\hline
Method & mIoU & aero & bag & cap & car & chair & EP & guitar & knife & lamp & laptop & motor & mug & pistol & rocket & skate & table\\
\hline
3DShapeNet \cite{yi2016scalable} & 81.4 & 81.0 & 78.4 & 77.7 & 75.7 & 87.6 & 61.9 & 92.0 & 85.4 & 82.5 & 95.7 & 70.6 & 91.9 & 85.9 & 53.1 & 69.8 & 75.3 \\
O-CNN \cite{wang2017cnn} & 85.9 & 85.5 & 87.1 & 84.7 & 77.0 & 91.1 & 85.1 & 91.9 & 87.4 & 83.3 & 95.4 & 56.9 & 96.2 & 81.6 & 53.5 & 74.1 & 84.4 \\
$\Psi$-CNN \cite{lei2019octree} & 86.8 & 84.2 & 82.1 & 83.8 & 80.5 & 91.0 & 78.3 & 91.6 & 86.7 & 84.7 & 95.6 & 74.8 & 94.5 & 83.4 & 61.3 & 75.9 & 85.9 \\
Kd-Net \cite{klokov2017escape} & 82.3 & 80.1 & 74.6 & 74.3 & 70.3 & 88.6 & 73.5 & 90.2 & 87.2 & 71.0 & 94.9 & 57.4 & 86.7 & 78.1 & 51.8 & 69.9 & 80.3 \\
3DmFV-Net \cite{Ben20183DmFV} & 84.3 & 82.0 & 84.3 & 86.0 & 76.9 & 89.9 & 73.9 & 90.8 & 85.7 & 82.6 & 95.2& 66.0 & 94.0 & 82.6 & 51.5& 73.5& 81.8 \\
ShapeContextNet \cite{xie2018attentional} & 84.6 & 83.8 & 80.8 & 83.5 & 79.3 & 90.5 & 69.8 & 91.7 & 86.5 & 82.9 & 96.0 & 69.2 & 93.8 & 82.5 & 62.9 & 74.4 & 80.8 \\
PointNet \cite{qi2017pointnet} & 83.7 & 83.4 & 78.7 & 82.5 & 74.9 & 89.6 & 73.0 & 91.5 & 85.9 & 80.8 & 95.3 & 65.2 & 93.0 & 81.2 & 57.9 & 72.8 & 80.6 \\
PointNet++ \cite{qi2017pointnet++} & 85.1 & 82.4 & 79.0 & 87.7 & 77.3 & 90.8 & 71.8 & 91.0 & 85.9 & 83.7 & 95.3 & 71.6 & 94.1 & 81.3 & 58.7 & 76.4 & 82.6\\
PCNN \cite{atzmon2018point} & 85.1 & 82.4 & 80.1 & 85.5 & 79.5 & 90.8 & 73.2 & 91.3 & 86.0 & 85.0 & 95.7 & 73.2 & 94.8 & 83.3 & 51.0 & 75.0 & 81.8 \\
SSCN \cite{graham20183d} & 86.0 & 84.1 & 83.0 & 84.0 & 80.8 & 91.4 & 78.2 & 91.6 & 89.1 & 85.0 & 95.8 & 73.7 & 95.2 & 84.0 & 58.5 & 76.0 & 82.7 \\
PointCNN \cite{li2018pointcnn} & 86.14 & 84.11 & 86.47 & 86.04 & 80.83 & 90.62 & 79.70 & 92.32 & 88.44 & 85.31 & 96.11 & 77.20 & 95.28 & 84.21 & 64.23 & 80.0 & 82.99  \\
SO-Net \cite{li2018so} & 84.9 & 82.8 & 77.8 & 88.0 & 77.3 & 90.6 & 73.5 & 90.7 & 83.9 & 82.8 & 94.8 & 69.1 & 94.2 & 80.9 & 53.1 & 72.9 & 83.0 \\
SPLATNet \cite{su2018splatnet} & 85.4 & 83.2 & 84.3 & 89.1& 80.3 & 90.7 & 75.5 & 92.1 & 87.1 & 83.9 & 96.3 & 75.6 & 95.8 & 83.8 & 64.0 & 75.5 & 81.8\\
SGPN \cite{wang2018sgpn} & 85.8 & 80.4 & 78.6 & 78.8 & 71.5 & 88.6 & 78.0 & 90.9 & 83.0 & 78.8 & 95.8 & 77.8 & 93.8 & 87.4 & 60.1 & 92.3& 89.4 \\
SpiderCNN \cite{xu2018spidercnn} & 85.3 & 83.5 & 81& 87.2 & 77.5& 90.7 & 76.8 & 91.1 & 87.3 & 83.3 & 95.8 & 70.2 & 93.5 & 82.7 & 59.7 & 75.8 & 82.8 \\
SRN \cite{duan2019structural} & 85.3 & 82.4 & 79.8 & 88.1 & 77.9 & 90.7 & 69.6 & 90.9 & 86.3 & 84.0 & 95.4 & 72.2 & 94.9 & 81.3 & 62.1 & 75.9 & 83.2 \\
RS-CNN \cite{liu2019relation}& 86.2 & 83.5 & 84.8 & 88.8 & 79.6 & 91.2 & 81.1 & 91.6 & 88.4 & 86.0 & 96.0 & 73.7 & 94.1 & 83.4 & 60.5 & 77.7 & 83.6\\
Point2Sequence \cite{liu2019point2sequence} & 85.2 & 82.6 & 81.8 & 87.5 & 77.3 & 90.8 & 77.1 & 91.1 & 86.9 & 83.9 & 95.7 & 70.8 & 94.6 & 79.3 & 58.1 & 75.2 & 82.8\\
SFCNN \cite{rao2019spherical} & 85.4 & 83.0 & 83.4 & 87.0 & 80.2 & 90.1 & 75.9 & 91.1 & 86.2 & 84.2 & 96.7 & 69.5 & 94.8 & 82.5 & 59.9 & 75.1 & 82.9 \\
SyncSpecCNN \cite{Li2017SyncSpecCNN}&84.7& 81.6 & 81.7 & 81.9 & 75.2 & 90.2 & 74.9 & 93.0 & 86.1 & 84.7 & 95.6 & 66.7 &92.7 & 81.6 & 60.6 & 82.9 & 82.1 \\
KC-Net \cite{shen2018mining} & 83.7 & 82.8 & 81.5 & 86.4 & 77.6 & 90.3 & 76.8 & 91.0 & 87.2 & 84.5 & 95.5 & 69.2 & 94.4 & 81.6 & 60.1 & 75.2 & 81.3 \\
RGCNN \cite{te2018rgcnn} & 84.3 & 80.2 & 82.8 & 92.6 & 75.3 & 89.2 & 73.7 & 91.3 & 88.4 & 83.3 & 96.0 & 63.9 & 95.7 & 60.9 & 44.6 & 72.9 & 80.4 \\
Spec-GCN \cite{wang2018local}& 85.4 & - & - & - & - & - & - & - & - & - & - & - & - & - & - & -\\
DGCNN \cite{wang2018dynamic} & 85.2 & 84.0 & 83.4 & 86.7 & 77.8 & 90.6 & 74.7 & 91.2 & 87.5 & 82.8 & 95.7 & 66.3 & 94.9 & 81.1 & 63.5 & 74.5 & 82.6\\
PointConv \cite{wu2019pointconv} & 85.7 & - & - & - & - & - & - & - & - & - & - & - & - & - & - & -\\
\hline
\end{tabular}}
\end{table*}

\begin{table*}[!t]
\caption{Experimental comparison of semantic segmentation with the state-of-the-art approaches on  SI3D}
\label{tab:SemanticSegmentation}
\centering
\resizebox{\textwidth}{!}{
\begin{tabular}{c|ccc|cccccccccccccc}
\hline
Method & OA & mAcc & mIoU & ceiling & floor & wall & beam & column & window & door & table & chair  & sofa & bookcase & board & clutter\\
\hline
SegCloud \cite{tchapmi2017segcloud} & - & 57.35 & 48.92 & 90.06 & 96.05 & 69.86 & 0.00 & 18.37 & 38.35 & 23.12 & 70.40 & 75.89 & 40.88 & 58.42 & 12.96 & 41.60 \\
PCCN \cite{wang2018deep} & - & 67.01 & 58.27 & 92.26 & 96.20 & 75.89 & 0.27 & 5.98 & 69.49 & 63.45 & 66.87 & 65.63 & 47.28 & 68.91 & 59.10 & 46.22 \\
ShapeContextNet \cite{xie2018attentional} &82.59 & - & 52.72 & -& -& -& -& -& -& -& -& -& -& -& -& -\\ 
PointNet \cite{qi2017pointnet} & - & 48.98 & 41.09 & 88.80 & 97.33 & 69.80 & 0.05 & 3.92 & 46.26 & 10.76 & 58.93 & 52.61 & 5.85 & 40.28 & 26.38 & 33.22 \\
PointCNN \cite{li2018pointcnn} & 85.91 & 63.86 & 57.26 & 92.31 & 98.24 & 79.41 & 0.00 & 17.60 & 22.77 & 62.09 & 74.39 & 80.59 & 31.67 & 66.67 & 62.05 & 56.74 \\
SGPN \cite{wang2018sgpn} & - & - & 54.35 & 79.44 & 66.29 & 88.77 & 77.98 & 60.71 & 66.62 & 56.75 & 46.90 & 40.77 & 6.38 & 47.61 & 11.05&-\\
SPGraph \cite{landrieu2018large}& 86.38 & 66.50 & 58.04 & 89.35 & 96.87 & 78.12 & 0.00 & 42.81 & 48.93 & 61.58 & 84.66 & 75.41 & 69.84 & 52.60 & 2.10 & 52.22 \\
DGCNN \cite{wang2018dynamic} &84.1&-&56.1 & -& -& -& -& -& -& -& -& -& -& -& -& -\\ 
Point2Note \cite{han2019point2node} & 89.01 & 79.10 & 70.00 & 94.08 & 97.28 & 83.42& 62.68 & 52.28 & 72.31 & 64.30 & 75.77 & 70.78 & 65.73 & 49.83 & 60.26 & 60.90 \\
HPEINet \cite{Jiang2019Hierarchical} & 87.18 & 68.30 & 61.85 & 91.47 & 98.16 & 81.38 & 0.00 & 23.34 & 65.30 & 40.02 & 75.46 & 87.70 & 58.45 & 67.78 & 65.61 & 49.36 \\
GACNet \cite{wang2019graph} & 87.79 & -& 62.85 & 92.28 & 98.27 & 81.90 & 0.00 & 20.35 & 59.07 & 40.85 & 78.54 & 85.80 & 61.70 & 70.75 & 74.66 & 52.82 \\ 
PointWeb \cite{zhao2019pointweb} & 86.97 & 66.64 & 60.28 & 91.95 & 98.48 & 79.39 & 0.00 & 21.11& 59.72 & 34.81 & 76.33 & 88.27 & 46.89 & 69.30 & 64.91 & 52.46\\

\hline
\end{tabular}}
\end{table*}

\section{Future Research Directions}
\label{future}
Despite remarkable development in this research area, either extraction of hand-crafted 3D point cloud descriptors or processing 3D point cloud using deep learning architecture is still reported as a challenging problem. According to the description about the contemporary works, we present a discussion of the future research directions.

\begin{enumerate}
    \item[$\bullet$] \hl{Nowadays, the 3D point cloud usually contains color or intensity attribute which provides much richer cues for point cloud processing. Therefore, a handful of studies} \cite{wang2018dynamic} \cite{xu2020weakly} \hl{have begun to fuse the geometric and  photometric information together to perform specific tasks. And the geometry-color or geometry-intensity information will be expect to improve the comprehensive performance of descriptors. }
        
    \item[$\bullet$] \hl{The limitations of 3D sensors, quantization artifact, reflective nature of the surface and unwanted objects inevitably lead to the presence of raw 3D point cloud with noise (including perturbation of points and outliers), occlusion as well as missing regions, which challenge the power of 3D point cloud descriptors. Therefore, how to address these problems for the improvement of the robustness of descriptors is still an open problem.} \hl{PointCleanNet}\cite{rakotosaona2020pointcleannet} and \hl{PointFilter}\cite{zhang2020pointfilter} \hl{reduce noise from the deep learning view of point. And the appearance of PF-Net} \cite{Huang2020PFNet} \hl{can perform completion for missing point cloud. These methods actually provide novel research ideas for settling these above issues.  }
    
    \item[$\bullet$] \hl{For deep learning based approaches, large amount of data has dramatically positive effect on the improvement of descriptiveness performance. Although few new datesets (e.g. Semantic KITTI} \cite{behley2019semantickitti:}, \hl{Oxford Radar Robotcar Dataset} \cite{RadarRobotCarDatasetArXiv}) \hl{have arisen, large point cloud datesets for specific tasks are actually rare at present. Therefore, it is desirable to solve the  the lack of training data is a crucial issue required to be solved. }
    
    \item[$\bullet$] \hl{Most of the current studies focus on 3D point cloud with fixed number of points (e.g. 2048 points) or fixed size (1m $\times$ 1m), which achieve state-of-the-art performance. However, the real-world point clouds are commonly large scale to which these methods cannot be extended. The proposals of PCT} \cite{chen2019pct} \hl{and RandLA-Net} \cite{hu2020randla} \hl{makes it possible to handle large-scale point cloud. Therefore, learning from large-scale point cloud is a promising direction for further exploration.}
    
    \item[$\bullet$] \hl{Joint 2D-3D reasoning would be an interesting research direction worth further investigation due to the fact that 3D point cloud as well as 2D image of the same object can be obtained at the same time in many practical applications. xMUDA} \cite{jaritz2020xmuda} \hl{exploits a cross-modal framework to learn 3D features for 3D semantic segmentation. } 

\end{enumerate}

\section{Conclusion}
\label{con}
 This paper specifically concentrates on the research activity concerning the field of 3D point cloud descriptors. We summary the main characteristics of the existing state-of-the-art 3D point cloud descriptors, ranging from early traditional hand-crafted descriptors to recent deep learning based descriptors. Advantages and disadvantages of various types of these descriptors are also analyzed. Although the rapid development of 3D point cloud descriptors has already gained promising results in many applications, it is believed that how to design a powerful descriptor further is still a challenging research area due to the complex environment, the presence of occlusion and clutter and other nuisances, the irregular nature of point cloud for deep learning.

\begin{acknowledgements}
\hl{This research was supported by the Fundamental Research Funds for the Central Universities (No. SWU120005)}
\end{acknowledgements}

%
%

\bibliographystyle{spmpsci}      
\bibliography{paper}   

\begin{thebibliography}{100}
\providecommand{\url}[1]{{#1}}
\providecommand{\urlprefix}{URL }
\expandafter\ifx\csname urlstyle\endcsname\relax
  \providecommand{\doi}[1]{DOI~\discretionary{}{}{}#1}\else
  \providecommand{\doi}{DOI~\discretionary{}{}{}\begingroup
  \urlstyle{rm}\Url}\fi

\bibitem{Aldoma2012Tutorial}
Aldoma, A., Marton, Z.C., Tombari, F., Wohlkinger, W., Potthast, C., Zeisl, B.,
  Rusu, R.B., Gedikli, S., Vincze, M.: Tutorial: Point cloud library:
  Three-dimensional object recognition and 6 dof pose estimation.
\newblock Robotics and Automation Magazine IEEE \textbf{19}(3), 80--91 (2012)

\bibitem{Aldoma2012OUR}
Aldoma, A., Tombari, F., Rusu, R.B., Vincze, M.: Our-cvfh – oriented, unique
  and repeatable clustered viewpoint feature histogram for object recognition
  and 6dof pose estimation.
\newblock In: Joint Dagm, pp. 113--122 (2012)

\bibitem{Aldoma2012CAD}
Aldoma, A., Vincze, M., Blodow, N., Gossow, D., Gedikli, S., Rusu, R.B.,
  Bradski, G.: Cad-model recognition and 6dof pose estimation using 3d cues.
\newblock In: IEEE International Conference on Computer Vision Workshops, pp.
  585--592 (2012)

\bibitem{Alexander2008Object}
Alexander~Patterson, I.V., Mordohai, P., Daniilidis, K.: Object Detection from
  Large-Scale 3D Datasets Using Bottom-Up and Top-Down Descriptors.
\newblock Springer Berlin Heidelberg (2008)

\bibitem{Alexandre20123D}
Alexandre, L.A.: 3d descriptors for object and category recognition: a
  comparative evaluation (2012)

\bibitem{Ali2014Contextual}
Ali, H., Shafait, F., Giannakidou, E., Vakali, A., Figueroa, N., Varvadoukas,
  T., Mavridis, N.: Contextual object category recognition for rgb-d scene
  labeling.
\newblock Robotics and Autonomous Systems \textbf{62}(2), 241--256 (2014)

\bibitem{Anand2013Contextually}
Anand, A., Koppula, H.S., Joachims, T., Saxena, A.: Contextually guided
  semantic labeling and search for three-dimensional point clouds.
\newblock International Journal of Robotics Research \textbf{32}(1), 19--34
  (2013)

\bibitem{Anguelov2005}
Anguelov, D., Taskarf, B., Chatalbashev, V., Koller, D., Gupta, D., Heitz, G.,
  Ng, A.: Discriminative learning of markov random fields for segmentation of
  3d scan data.
\newblock In: 2005 IEEE Computer Society Conference on Computer Vision and
  Pattern Recognition (CVPR'05), vol.~2, pp. 169--176 vol. 2 (2005).
\newblock \doi{10.1109/CVPR.2005.133}

\bibitem{2017arXiv170201105A}
{Armeni}, I., {Sax}, A., {Zamir}, A.R., {Savarese}, S.: {Joint 2D-3D-Semantic
  Data for Indoor Scene Understanding}.
\newblock ArXiv e-prints  (2017)

\bibitem{armeni20163d}
Armeni, I., Sener, O., Zamir, A.R., Jiang, H., Brilakis, I., Fischer, M.,
  Savarese, S.: 3d semantic parsing of large-scale indoor spaces.
\newblock In: Computer Vision and Pattern Recognition (2016)

\bibitem{atzmon2018point}
Atzmon, M., Maron, H., Lipman, Y.: Point convolutional neural networks by
  extension operators.
\newblock arXiv preprint arXiv:1803.10091  (2018)

\bibitem{RadarRobotCarDatasetArXiv}
Barnes, D., Gadd, M., Murcutt, P., Newman, P., Posner, I.: The oxford radar
  robotcar dataset: A radar extension to the oxford robotcar dataset.
\newblock arXiv preprint arXiv: 1909.01300  (2019).
\newblock \urlprefix\url{https://arxiv.org/pdf/1909.01300}

\bibitem{behley2019semantickitti:}
Behley, J., Garbade, M., Milioto, A., Quenzel, J., Behnke, S., Stachniss, C.,
  Gall, J.: Semantickitti: A dataset for semantic scene understanding of lidar
  sequences pp. 9297--9307 (2019)

\bibitem{Behley2012}
Behley, J., Steinhage, V., Cremers, A.B.: Performance of histogram descriptors
  for the classification of 3d laser range data in urban environments.
\newblock In: 2012 IEEE International Conference on Robotics and Automation,
  pp. 4391--4398 (2012).
\newblock \doi{10.1109/ICRA.2012.6225003}

\bibitem{7139443}
Beksi, W.J., Papanikolopoulos, N.: Object classification using dictionary
  learning and rgb-d covariance descriptors.
\newblock In: 2015 IEEE International Conference on Robotics and Automation
  (ICRA), pp. 1880--1885 (2015).
\newblock \doi{10.1109/ICRA.2015.7139443}

\bibitem{Ben20183DmFV}
Ben-Shabat, Y., Lindenbaum, M., Fischer, A.: 3dmfv: 3d point cloud
  classification in real-time using convolutional neural network.
\newblock IEEE Robotics and Automation Letters \textbf{PP}(99), 1--1 (2018)

\bibitem{Beserra2013Efficient}
Beserra~Gomes, R., Marques Ferreira Da~Silva, B., Rocha, L.K.D.M., Aroca, R.V.,
  Gon, Alves, L.M.G.: Efficient 3d object recognition using foveated point
  clouds.
\newblock Computers and Graphics \textbf{37}(5), 496--508 (2013)

\bibitem{Bo2011Depth}
Bo, L., Ren, X., Fox, D.: Depth kernel descriptors for object recognition
  \textbf{32}(14), 821--826 (2011)

\bibitem{Chan2014A}
Chan, K.C., Koh, C.K., Lee, C.S.G.: A 3-d-point-cloud system for human-pose
  estimation.
\newblock IEEE Transactions on Systems Man and Cybernetics Systems
  \textbf{44}(11), 1486--1497 (2014)

\bibitem{Chang2015ShapeNet}
Chang, A.X., Funkhouser, T., Guibas, L., Hanrahan, P., Huang, Q., Li, Z.,
  Savarese, S., Savva, M., Song, S., Su, H.: Shapenet: An information-rich 3d
  model repository.
\newblock Computer Science  (2015)

\bibitem{Charles2017PointNet}
Charles, R.Q., Hao, S., Mo, K., Guibas, L.J.: Pointnet: Deep learning on point
  sets for 3d classification and segmentation.
\newblock In: IEEE Conference on Computer Vision and Pattern Recognition (2017)

\bibitem{chatfield2014return}
Chatfield, K., Simonyan, K., Vedaldi, A., Zisserman, A.: Return of the devil in
  the details: Delving deep into convolutional nets.
\newblock arXiv preprint arXiv:1405.3531  (2014)

\bibitem{chen2019pct}
Chen, S., Niu, S., Lan, T., Liu, B.: Pct: Large-scale 3d point cloud
  representations via graph inception networks with applications to autonomous
  driving.
\newblock In: 2019 IEEE International Conference on Image Processing (ICIP),
  pp. 4395--4399. IEEE (2019)

\bibitem{Chen2014Performance}
Chen, T., Dai, B., Liu, D., Song, J.: Performance of global descriptors for
  velodyne-based urban object recognition.
\newblock IEEE (2014)

\bibitem{chen2017multi}
Chen, X., Ma, H., Wan, J., Li, B., Xia, T.: Multi-view 3d object detection
  network for autonomous driving.
\newblock In: Proceedings of the IEEE Conference on Computer Vision and Pattern
  Recognition, pp. 1907--1915 (2017)

\bibitem{Cheng2014Robust}
Cheng, J., Xiang, Z., Cao, T., Liu, J.: Robust vehicle detection using 3d lidar
  under complex urban environment.
\newblock In: IEEE International Conference on Robotics and Automation, pp.
  691--696 (2014)

\bibitem{Cirujeda2014MCOV}
Cirujeda, P., Mateo, X., Dicente, Y., Binefa, X.: Mcov: A covariance descriptor
  for fusion of texture and shape features in 3d point clouds.
\newblock In: International Conference on 3d Vision, pp. 551--558 (2014)

\bibitem{dai2017scannet}
Dai, A., Chang, A.X., Savva, M., Halber, M., Funkhouser, T., Nie{\ss}ner, M.:
  Scannet: Richly-annotated 3d reconstructions of indoor scenes.
\newblock In: Proceedings of the IEEE Conference on Computer Vision and Pattern
  Recognition, pp. 5828--5839 (2017)

\bibitem{dai20183dmv}
Dai, A., Nie{\ss}ner, M.: 3dmv: Joint 3d-multi-view prediction for 3d semantic
  scene segmentation.
\newblock In: Proceedings of the European Conference on Computer Vision (ECCV),
  pp. 452--468 (2018)

\bibitem{de2013unsupervised}
De~Deuge, M., Quadros, A., Hung, C., Douillard, B.: Unsupervised feature
  learning for classification of outdoor 3d scans.
\newblock In: Australasian Conference on Robitics and Automation, vol.~2, p.~1
  (2013)

\bibitem{deng2018ppf}
Deng, H., Birdal, T., Ilic, S.: Ppf-foldnet: Unsupervised learning of rotation
  invariant 3d local descriptors.
\newblock In: Proceedings of the European Conference on Computer Vision (ECCV),
  pp. 602--618 (2018)

\bibitem{deng2018ppfnet}
Deng, H., Birdal, T., Ilic, S.: Ppfnet: Global context aware local features for
  robust 3d point matching.
\newblock In: Proceedings of the IEEE Conference on Computer Vision and Pattern
  Recognition, pp. 195--205 (2018)

\bibitem{Drost2010Model}
Drost, B., Ulrich, M., Navab, N., Ilic, S.: Model globally, match locally:
  Efficient and robust 3d object recognition.
\newblock In: Computer Vision and Pattern Recognition, pp. 998--1005 (2010)

\bibitem{duan2019structural}
Duan, Y., Zheng, Y., Lu, J., Zhou, J., Tian, Q.: Structural relational
  reasoning of point clouds.
\newblock In: Proceedings of the IEEE Conference on Computer Vision and Pattern
  Recognition, pp. 949--958 (2019)

\bibitem{engelcke2017vote3deep}
Engelcke, M., Rao, D., Wang, D.Z., Tong, C.H., Posner, I.: Vote3deep: Fast
  object detection in 3d point clouds using efficient convolutional neural
  networks.
\newblock In: 2017 IEEE International Conference on Robotics and Automation
  (ICRA), pp. 1355--1361. IEEE (2017)

\bibitem{Fehr2014RGB}
Fehr, D., Beksi, W.J., Zermas, D., Papanikolopoulos, N.: Rgb-d object
  classification using covariance descriptors.
\newblock In: IEEE International Conference on Robotics and Automation, pp.
  5467--5472 (2014)

\bibitem{Fehr2012Compact}
Fehr, D., Cherian, A., Sivalingam, R., Nickolay, S.: Compact covariance
  descriptors in 3d point clouds for object recognition.
\newblock In: IEEE International Conference on Robotics and Automation, pp.
  1793--1798 (2012)

\bibitem{feng2018gvcnn}
Feng, Y., Zhang, Z., Zhao, X., Ji, R., Gao, Y.: Gvcnn: Group-view convolutional
  neural networks for 3d shape recognition.
\newblock In: Proceedings of the IEEE Conference on Computer Vision and Pattern
  Recognition, pp. 264--272 (2018)

\bibitem{Fiolka2012SURE}
Fiolka, T., Stückler, J., Klein, D.A., Schulz, D., Behnke, S.: Sure: Surface
  entropy for distinctive 3d features.
\newblock In: International Conference on Spatial Cognition, pp. 74--93 (2012)

\bibitem{Flint2007Thrift}
Flint, A., Dick, A., Hengel, A.V.D.: Thrift: Local 3d structure recognition.
\newblock In: Digital Image Computing Techniques and Applications, Biennial
  Conference of the Australian Pattern Recognition Society on, pp. 182--188
  (2007)

\bibitem{Frome2004}
Frome, A., Huber, D., Kolluri, R., B{\"u}low, T., Malik, J.: Recognizing
  Objects in Range Data Using Regional Point Descriptors, pp. 224--237.
\newblock Springer Berlin Heidelberg, Berlin, Heidelberg (2004)

\bibitem{gadelha2018multiresolution}
Gadelha, M., Wang, R., Maji, S.: Multiresolution tree networks for 3d point
  cloud processing.
\newblock In: Proceedings of the European Conference on Computer Vision (ECCV),
  pp. 103--118 (2018)

\bibitem{Ge2016Non}
Ge, X.: Non-rigid registration of 3d point clouds under isometric deformation.
\newblock Isprs Journal of Photogrammetry and Remote Sensing \textbf{121},
  192--202 (2016)

\bibitem{geiger2012we}
Geiger, A., Lenz, P., Urtasun, R.: Are we ready for autonomous driving? the
  kitti vision benchmark suite.
\newblock In: 2012 IEEE Conference on Computer Vision and Pattern Recognition,
  pp. 3354--3361. IEEE (2012)

\bibitem{Girshick2015Fast}
Girshick, R.: Fast r-cnn.
\newblock In: IEEE International Conference on Computer Vision (2015)

\bibitem{Golovinskiy2009}
Golovinskiy, A., Kim, V.G., Funkhouser, T.: Shape-based recognition of 3d point
  clouds in urban environments.
\newblock In: 2009 IEEE 12th International Conference on Computer Vision, pp.
  2154--2161 (2009)

\bibitem{graham20183d}
Graham, B., Engelcke, M., van~der Maaten, L.: 3d semantic segmentation with
  submanifold sparse convolutional networks.
\newblock In: Proceedings of the IEEE Conference on Computer Vision and Pattern
  Recognition, pp. 9224--9232 (2018)

\bibitem{guinard2017weakly}
Guinard, S., Landrieu, L.: Weakly supervised segmentation-aided classification
  of urban scenes from 3d lidar point clouds.
\newblock In: ISPRS Workshop 2017 (2017)

\bibitem{Guo2016A}
Guo, Y., Bennamoun, M., Sohel, F., Lu, M., Wan, J., Kwok, N.M.: A comprehensive
  performance evaluation of 3d local feature descriptors.
\newblock International Journal of Computer Vision \textbf{116}(1), 66--89
  (2016)

\bibitem{Guo2013Rotational}
Guo, Y., Sohel, F., Bennamoun, M., Lu, M., Wan, J.: Rotational projection
  statistics for 3d local surface description and object recognition.
\newblock International Journal of Computer Vision \textbf{105}(1), 63--86
  (2013)

\bibitem{hackel2017isprs}
Hackel, T., Savinov, N., Ladicky, L., Wegner, J.D., Schindler, K., Pollefeys,
  M.: Semantic3d.net: A new large-scale point cloud classification benchmark.
\newblock In: ISPRS Annals of the Photogrammetry, Remote Sensing and Spatial
  Information Sciences, vol. IV-1-W1, pp. 91--98 (2017)

\bibitem{Hadji2014Local}
Hadji, I., Desouza, G.N.: Local-to-global signature descriptor for 3d object
  recognition.
\newblock In: Asian Conference on Computer Vision, pp. 570--584 (2014)

\bibitem{han2019point2node}
Han, W., Wen, C., Wang, C., Li, X., Li, Q.: Point2node: Correlation learning of
  dynamic-node for point cloud feature modeling.
\newblock arXiv preprint arXiv:1912.10775  (2019)

\bibitem{han2015matchnet}
Han, X., Leung, T., Jia, Y., Sukthankar, R., Berg, A.C.: Matchnet: Unifying
  feature and metric learning for patch-based matching.
\newblock In: Proceedings of the IEEE Conference on Computer Vision and Pattern
  Recognition, pp. 3279--3286 (2015)

\bibitem{Han2017A}
Han, X.F., Jin, J.S., Wang, M.J., Jiang, W., Gao, L., Xiao, L.: A review of
  algorithms for filtering the 3d point cloud.
\newblock Signal Processing Image Communication \textbf{57}, 103--112 (2017)

\bibitem{He2014Spatial}
He, K., Zhang, X., Ren, S., Sun, J.: Spatial pyramid pooling in deep
  convolutional networks for visual recognition.
\newblock IEEE Transactions on Pattern Analysis and Machine Intelligence
  \textbf{37}(9), 1904--16 (2014)

\bibitem{Himmelsbach2009Real}
Himmelsbach, M., Luettel, T., Wuensche, H.J.: Real-time object classification
  in 3d point clouds using point feature histograms.
\newblock In: Ieee/rsj International Conference on Intelligent Robots and
  Systems, pp. 994--1000 (2009)

\bibitem{hu2020randla}
Hu, Q., Yang, B., Xie, L., Rosa, S., Guo, Y., Wang, Z., Trigoni, N., Markham,
  A.: Randla-net: Efficient semantic segmentation of large-scale point clouds.
\newblock In: Proceedings of the IEEE/CVF Conference on Computer Vision and
  Pattern Recognition, pp. 11108--11117 (2020)

\bibitem{Huang2012}
Huang, J., You, S.: Point cloud matching based on 3d self-similarity.
\newblock In: 2012 IEEE Computer Society Conference on Computer Vision and
  Pattern Recognition Workshops, pp. 41--48 (2012).
\newblock \doi{10.1109/CVPRW.2012.6238913}

\bibitem{Huang2013}
Huang, J., You, S.: Detecting objects in scene point cloud: A combinational
  approach.
\newblock In: 2013 International Conference on 3D Vision - 3DV 2013, pp.
  175--182 (2013).
\newblock \doi{10.1109/3DV.2013.31}

\bibitem{huang2018recurrent}
Huang, Q., Wang, W., Neumann, U.: Recurrent slice networks for 3d segmentation
  of point clouds.
\newblock In: Proceedings of the IEEE Conference on Computer Vision and Pattern
  Recognition, pp. 2626--2635 (2018)

\bibitem{Huang2020PFNet}
Huang, Z., Yu, Y., Xu, J., Ni, F., Le, X.: Pf-net: Point fractal network for 3d
  point cloud completion.
\newblock In: Proceedings of the IEEE/CVF Conference on Computer Vision and
  Pattern Recognition, pp. 7662--7670 (2020)

\bibitem{huang2019deepccfv}
Huang, Z., Zhao, Z., Zhou, H., Zhao, X., Gao, Y.: Deepccfv: Camera
  constraint-free multi-view convolutional neural network for 3d object
  retrieval  (2019)

\bibitem{Hwang2012Robust}
Hwang, H., Hyung, S., Yoon, S., Roh, K.: Robust descriptors for 3d point clouds
  using geometric and photometric local feature.
\newblock In: Ieee/rsj International Conference on Intelligent Robots and
  Systems, pp. 4027--4033 (2012)

\bibitem{Ip2002Using}
Ip, C.Y., Lapadat, D., Sieger, L., Regli, W.C.: Using shape distributions to
  compare solid models.
\newblock In: ACM Symposium on Solid Modeling and Applications, pp. 273--280
  (2002)

\bibitem{jaritz2019multi}
Jaritz, M., Gu, J., Su, H.: Multi-view pointnet for 3d scene understanding.
\newblock In: Proceedings of the IEEE International Conference on Computer
  Vision Workshops, pp. 0--0 (2019)

\bibitem{jaritz2020xmuda}
Jaritz, M., Vu, T.H., Charette, R.d., Wirbel, E., P{\'e}rez, P.: xmuda:
  Cross-modal unsupervised domain adaptation for 3d semantic segmentation.
\newblock In: Proceedings of the IEEE/CVF Conference on Computer Vision and
  Pattern Recognition, pp. 12605--12614 (2020)

\bibitem{Jiang2019Hierarchical}
Jiang, L., Zhao, H., Liu, S., Shen, X., Jia, J.: Hierarchical point-edge
  interaction network for point cloud semantic segmentation  (2019)

\bibitem{Johnson1998Surface}
Johnson, A.E., Hebert, M.: Surface matching for object recognition in complex
  3-d scenes.
\newblock Image and Vision Computing \textbf{16}(9-10), 635--651 (1998)

\bibitem{765655}
Johnson, A.E., Hebert, M.: Using spin images for efficient object recognition
  in cluttered 3d scenes.
\newblock IEEE Transactions on Pattern Analysis and Machine Intelligence
  \textbf{21}(5), 433--449 (1999).
\newblock \doi{10.1109/34.765655}

\bibitem{Kahler2013Efficient}
Kahler, O., Reid, I.: Efficient 3d scene labeling using fields of trees.
\newblock In: IEEE International Conference on Computer Vision, pp. 3064--3071
  (2013)

\bibitem{Kasaei2016GOOD}
Kasaei, S.H., Tomé, A.M., Lopes, L.S., Oliveira, M.: Good: A global
  orthographic object descriptor for 3d object recognition and manipulation.
\newblock Pattern Recognition Letters \textbf{83}, 312--320 (2016)

\bibitem{khoury2017learning}
Khoury, M., Zhou, Q.Y., Koltun, V.: Learning compact geometric features.
\newblock In: Proceedings of the IEEE International Conference on Computer
  Vision, pp. 153--161 (2017)

\bibitem{klokov2017escape}
Klokov, R., Lempitsky, V.: Escape from cells: Deep kd-networks for the
  recognition of 3d point cloud models.
\newblock In: Proceedings of the IEEE International Conference on Computer
  Vision, pp. 863--872 (2017)

\bibitem{Lalonde2006Natural}
Lalonde, J.F., Vandapel, N., Huber, D.F., Hebert, M.: Natural terrain
  classification using three‐dimensional ladar data for ground robot
  mobility.
\newblock Journal of Field Robotics \textbf{23}(10), 839–861 (2006)

\bibitem{landrieu2018large}
Landrieu, L., Simonovsky, M.: Large-scale point cloud semantic segmentation
  with superpoint graphs.
\newblock In: Proceedings of the IEEE Conference on Computer Vision and Pattern
  Recognition, pp. 4558--4567 (2018)

\bibitem{Lehtom2016Object}
Lehtomäki, M., Jaakkola, A., Hyyppä, J., Lampinen, J., Kaartinen, H., Kukko,
  A., Puttonen, E., Hyyppä, H.: Object classification and recognition from
  mobile laser scanning point clouds in a road environment.
\newblock IEEE Transactions on Geoscience and Remote Sensing \textbf{54}(2),
  1226--1239 (2016)

\bibitem{lei2019octree}
Lei, H., Akhtar, N., Mian, A.: Octree guided cnn with spherical kernels for 3d
  point clouds.
\newblock In: Proceedings of the IEEE Conference on Computer Vision and Pattern
  Recognition (2019)

\bibitem{li2018so}
Li, J., Chen, B.M., Hee~Lee, G.: So-net: Self-organizing network for point
  cloud analysis.
\newblock In: Proceedings of the IEEE conference on computer vision and pattern
  recognition, pp. 9397--9406 (2018)

\bibitem{Li2016RISAS}
Li, X., Wu, K., Liu, Y., Ranasinghe, R., Dissanayake, G., Xiong, R.: Risas: A
  novel rotation, illumination, scale invariant appearance and shape feature
  (2016)

\bibitem{li2018pointcnn}
Li, Y., Bu, R., Sun, M., Wu, W., Di, X., Chen, B.: Pointcnn: Convolution on
  x-transformed points.
\newblock In: Advances in Neural Information Processing Systems, pp. 820--830
  (2018)

\bibitem{Li2017SyncSpecCNN}
Li, Y., Hao, S., Guo, X., Guibas, L.: Syncspeccnn: Synchronized spectral cnn
  for 3d shape segmentation.
\newblock In: 2017 IEEE Conference on Computer Vision and Pattern Recognition
  (CVPR) (2017)

\bibitem{Lin2017S}
Lin, B., Wang, F., Zhao, F., Sun, Y.: Scale invariant point feature (sipf) for
  3d point clouds and 3d multi-scale object detection.
\newblock Neural Computing and Applications  (2017).
\newblock \doi{10.1007/s00521-017-2964-1}

\bibitem{liu2019point2sequence}
Liu, X., Han, Z., Liu, Y.S., Zwicker, M.: Point2sequence: Learning the shape
  representation of 3d point clouds with an attention-based sequence to
  sequence network.
\newblock In: AAAI (2019)

\bibitem{liu2019relation}
Liu, Y., Fan, B., Xiang, S., Pan, C.: Relation-shape convolutional neural
  network for point cloud analysis.
\newblock In: Proceedings of the IEEE Conference on Computer Vision and Pattern
  Recognition (2019)

\bibitem{Logoglu2016CoSPAIR}
Logoglu, K.B., Kalkan, S., Temizel, A.: Cospair: Colored histograms of spatial
  concentric surflet-pairs for 3d object recognition.
\newblock Robotics and Autonomous Systems \textbf{75}(part B), 558--570 (2016)

\bibitem{Lima2016}
d.~M.~Lima, J.P.S., Teichrieb, V.: An efficient global point cloud descriptor
  for object recognition and pose estimation.
\newblock In: 2016 29th SIBGRAPI Conference on Graphics, Patterns and Images
  (SIBGRAPI), pp. 56--63 (2016).
\newblock \doi{10.1109/SIBGRAPI.2016.017}

\bibitem{Madry2012Improving}
Madry, M., Ek, C.H., Detry, R., Hang, K.: Improving generalization for 3d
  object categorization with global structure histograms.
\newblock In: Ieee/rsj International Conference on Intelligent Robots and
  Systems, pp. 1379--1386 (2012)

\bibitem{Marianna2012From}
Madry, M., Song, D., Kragic, D.: From object categories to grasp transfer using
  probabilistic reasoning.
\newblock Proceedings - IEEE International Conference on Robotics and
  Automation pp. 1716--1723 (2012)

\bibitem{Marton2011Combined}
Marton, Z.C., Pangercic, D., Blodow, N., Beetz, M.: Combined 2d–3d
  categorization and classification for multimodal perception systems.
\newblock International Journal of Robotics Research \textbf{30}(11),
  1378--1402 (2011)

\bibitem{Marton2010General}
Marton, Z.C., Pangercic, D., Blodow, N., Kleinehellefort, J., Beetz, M.:
  General 3d modelling of novel objects from a single view.
\newblock In: Ieee/rsj International Conference on Intelligent Robots and
  Systems, pp. 3700--3705 (2010)

\bibitem{Matei2006}
Matei, B., Shan, Y., Sawhney, H.S., Tan, Y., Kumar, R., Huber, D., Hebert, M.:
  Rapid object indexing using locality sensitive hashing and joint 3d-signature
  space estimation.
\newblock IEEE Transactions on Pattern Analysis and Machine Intelligence
  \textbf{28}(7), 1111--1126 (2006)

\bibitem{Mateo2014A}
Mateo, C.M., Gil, P., Torres, F.: A performance evaluation of surface
  normals-based descriptors for recognition of objects using cad-models.
\newblock In: International Conference on Informatics in Control, Automation
  and Robotics, pp. 428 -- 435 (2014)

\bibitem{maturana2015VoxNet}
Maturana, D., Scherer, S.: Voxnet: A 3d convolutional neural network for
  real-time object recognition.
\newblock In: 2015 IEEE/RSJ International Conference on Intelligent Robots and
  Systems (IROS), pp. 922--928. IEEE (2015)

\bibitem{Muja2011REIN}
Muja, M., Rusu, R.B., Bradski, G., Lowe, D.G.: Rein - a fast, robust, scalable
  recognition infrastructure pp. 2939--2946 (2011)

\bibitem{Munoz2012Co}
Munoz, D., Bagnell, J.A., Hebert, M.: Co-inference for multi-modal scene
  analysis.
\newblock In: Proceedings of the 12th European conference on Computer Vision -
  Volume Part VI (2012)

\bibitem{Munoz-2009-10227}
Munoz, D., Bagnell, J.A.D., Vandapel, N., Hebert, M.: Contextual classification
  with functional max-margin markov networks.
\newblock In: Proceedings of IEEE Computer Society Conference on Computer
  Vision and Pattern Recognition (CVPR) (2009)

\bibitem{phong1975illumination}
Phong, B.T.: Illumination for computer generated pictures.
\newblock Communications of the ACM \textbf{18}(6), 311--317 (1975)

\bibitem{Prakhya20173DHoPD}
Prakhya, S.M., Lin, J., Chandrasekhar, V., Lin, W., Liu, B.: 3dhopd: A fast
  low-dimensional 3-d descriptor.
\newblock IEEE Robotics and Automation Letters \textbf{2}(3), 1472--1479 (2017)

\bibitem{Prakhya2015B}
Prakhya, S.M., Liu, B., Lin, W.: B-shot: A binary feature descriptor for fast
  and efficient keypoint matching on 3d point clouds.
\newblock In: Ieee/rsj International Conference on Intelligent Robots and
  Systems, pp. 1929--1934 (2015)

\bibitem{qi2017pointnet}
Qi, C.R., Su, H., Mo, K., Guibas, L.J.: Pointnet: Deep learning on point sets
  for 3d classification and segmentation.
\newblock In: Proceedings of the IEEE Conference on Computer Vision and Pattern
  Recognition, pp. 652--660 (2017)

\bibitem{qi2016volumetric}
{Qi}, C.R., {Su}, H., {Nießner}, M., {Dai}, A., {Yan}, M., {Guibas}, L.J.:
  Volumetric and multi-view cnns for object classification on 3d data.
\newblock In: 2016 IEEE Conference on Computer Vision and Pattern Recognition
  (CVPR), pp. 5648--5656 (2016)

\bibitem{qi2017pointnet++}
Qi, C.R., Yi, L., Su, H., Guibas, L.J.: Pointnet++: Deep hierarchical feature
  learning on point sets in a metric space.
\newblock In: Advances in neural information processing systems, pp. 5099--5108
  (2017)

\bibitem{Radu2010NARF}
Radu, B.S., Rusu, B., Konolige, K., Burgard, W.: Narf: 3d range image features
  for object recognition  (2010)

\bibitem{Rahmani2014HOPC}
Rahmani, H., Mahmood, A., Du, Q.H., Mian, A.: Hopc: Histogram of oriented
  principal components of 3d pointclouds for action recognition.
\newblock In: European Conference on Computer Vision, pp. 742--757 (2014)

\bibitem{rakotosaona2020pointcleannet}
Rakotosaona, M.J., La~Barbera, V., Guerrero, P., Mitra, N.J., Ovsjanikov, M.:
  Pointcleannet: Learning to denoise and remove outliers from dense point
  clouds.
\newblock In: Computer Graphics Forum, vol.~39, pp. 185--203. Wiley Online
  Library (2020)

\bibitem{rao2019spherical}
Rao, Y., Lu, J., Zhou, J.: Spherical fractal convolutional neural networks for
  point cloud recognition.
\newblock In: Proceedings of the IEEE Conference on Computer Vision and Pattern
  Recognition, pp. 452--460 (2019)

\bibitem{ravanbakhsh2016deep}
Ravanbakhsh, S., Schneider, J., Poczos, B.: Deep learning with sets and point
  clouds.
\newblock arXiv preprint arXiv:1611.04500  (2016)

\bibitem{ren2015faster}
Ren, S., He, K., Girshick, R., Sun, J.: Faster r-cnn: Towards real-time object
  detection with region proposal networks.
\newblock In: Advances in neural information processing systems, pp. 91--99
  (2015)

\bibitem{riegler2017octnet}
Riegler, G., Osman~Ulusoy, A., Geiger, A.: Octnet: Learning deep 3d
  representations at high resolutions.
\newblock In: Proceedings of the IEEE Conference on Computer Vision and Pattern
  Recognition, pp. 3577--3586 (2017)

\bibitem{rostami2019a}
Rostami, R., Bashiri, F.S., Rostami, B., Yu, Z.: A survey on data‐driven 3d
  shape descriptors.
\newblock Computer Graphics Forum \textbf{38}(1), 356--393 (2019)

\bibitem{Xavier2018Paris}
Roynard, X., Deschaud, J.E., Goulette, F.: Paris-lille-3d: A point cloud
  dataset for urban scene segmentation and classification.
\newblock In: IEEE/CVF Conference on Computer Vision and Pattern Recognition
  Workshops (2018)

\bibitem{Rusu2009Fast}
Rusu, R.B., Blodow, N., Beetz, M.: Fast point feature histograms (fpfh) for 3d
  registration.
\newblock In: IEEE International Conference on Robotics and Automation, pp.
  1848--1853 (2009)

\bibitem{Rusu2008Aligning}
Rusu, R.B., Blodow, N., Marton, Z.C., Beetz, M.: Aligning point cloud views
  using persistent feature histograms.
\newblock In: Ieee/rsj International Conference on Intelligent Robots and
  Systems, pp. 3384--3391 (2008)

\bibitem{Rusu2014Fast}
Rusu, R.B., Bradski, G., Thibaux, R., Hsu, J.: Fast 3d recognition and pose
  using the viewpoint feature histogram.
\newblock In: Ieee/rsj International Conference on Intelligent Robots and
  Systems, pp. 2155--2162 (2014)

\bibitem{Rusu20113D}
Rusu, R.B., Cousins, S.: 3d is here: Point cloud library (pcl).
\newblock In: IEEE International Conference on Robotics and Automation, pp.
  1--4 (2011)

\bibitem{Rusu2009Detecting}
Rusu, R.B., Holzbach, A., Beetz, M., Bradski, G.: Detecting and segmenting
  objects for mobile manipulation.
\newblock In: IEEE International Conference on Computer Vision Workshops, pp.
  47--54 (2009)

\bibitem{Rusu2008}
Rusu, R.B., Marton, Z.C., Blodow, N., Beetz, M., Systems, I.A., München, T.U.:
  Persistent point feature histograms for 3d point clouds.
\newblock In: In Proceedings of the 10th International Conference on
  Intelligent Autonomous Systems (IAS-10 (2008)

\bibitem{Salti2012On}
Salti~S Petrelli~A, T.F.: On the affinity between 3d detectors and descriptors.
\newblock In: 3DIMPVT, pp. 424--431 (2012)

\bibitem{sanchez2013image}
S{\'a}nchez, J., Perronnin, F., Mensink, T., Verbeek, J.: Image classification
  with the fisher vector: Theory and practice.
\newblock International journal of computer vision \textbf{105}(3), 222--245
  (2013)

\bibitem{Shan2006}
Shan, Y., Sawhney, H.S., Matei, B., Kumar, R.: Shapeme histogram projection and
  matching for partial object recognition.
\newblock IEEE Transactions on Pattern Analysis and Machine Intelligence
  \textbf{28}(4), 568--577 (2006).
\newblock \doi{10.1109/TPAMI.2006.83}

\bibitem{Shelhamer2014Fully}
Shelhamer, E., Long, J., Darrell, T.: Fully convolutional networks for semantic
  segmentation.
\newblock IEEE Transactions on Pattern Analysis and Machine Intelligence
  \textbf{39}(4), 640--651 (2014)

\bibitem{shen2018mining}
Shen, Y., Feng, C., Yang, Y., Tian, D.: Mining point cloud local structures by
  kernel correlation and graph pooling.
\newblock In: Proceedings of the IEEE conference on computer vision and pattern
  recognition, pp. 4548--4557 (2018)

\bibitem{simonovsky2017dynamic}
Simonovsky, M., Komodakis, N.: Dynamic edge-conditioned filters in
  convolutional neural networks on graphs.
\newblock In: Proceedings of the IEEE conference on computer vision and pattern
  recognition, pp. 3693--3702 (2017)

\bibitem{singh2014bigbird}
Singh, A., Sha, J., Narayan, K.S., Achim, T., Abbeel, P.: Bigbird: A
  large-scale 3d database of object instances.
\newblock In: 2014 IEEE International Conference on Robotics and Automation
  (ICRA), pp. 509--516. IEEE (2014)

\bibitem{su2018splatnet}
Su, H., Jampani, V., Sun, D., Maji, S., Kalogerakis, E., Yang, M.H., Kautz, J.:
  Splatnet: Sparse lattice networks for point cloud processing.
\newblock In: Proceedings of the IEEE Conference on Computer Vision and Pattern
  Recognition, pp. 2530--2539 (2018)

\bibitem{su2015multi}
Su, H., Maji, S., Kalogerakis, E., Learned-Miller, E.: Multi-view convolutional
  neural networks for 3d shape recognition.
\newblock In: Proceedings of the IEEE international conference on computer
  vision, pp. 945--953 (2015)

\bibitem{szegedy2016rethinking}
Szegedy, C., Vanhoucke, V., Ioffe, S., Shlens, J., Wojna, Z.: Rethinking the
  inception architecture for computer vision.
\newblock In: Proceedings of the IEEE conference on computer vision and pattern
  recognition, pp. 2818--2826 (2016)

\bibitem{Tang2016Signature}
Tang, K., Song, P., Chen, X.: Signature of geometric centroids for 3d local
  shape description and partial shape matching.
\newblock In: Asian Conference on Computer Vision, pp. 311--326 (2016)

\bibitem{tchapmi2017segcloud}
Tchapmi, L., Choy, C., Armeni, I., Gwak, J., Savarese, S.: Segcloud: Semantic
  segmentation of 3d point clouds.
\newblock In: 2017 International Conference on 3D Vision (3DV), pp. 537--547.
  IEEE (2017)

\bibitem{te2018rgcnn}
Te, G., Hu, W., Zheng, A., Guo, Z.: Rgcnn: Regularized graph cnn for point
  cloud segmentation.
\newblock In: 2018 ACM Multimedia Conference on Multimedia Conference, pp.
  746--754. ACM (2018)

\bibitem{Tombari2010Uniqueshape}
Tombari, F., Salti, S., Stefano, L.D.: Unique shape context for 3d data
  description.
\newblock In: Proceedings of the ACM workshop on 3D object retrieval, pp.
  57--62 (2010)

\bibitem{Tombari2010Unique}
Tombari, F., Salti, S., Stefano, L.D.: Unique signatures of histograms for
  local surface description.
\newblock In: European Conference on Computer Vision Conference on Computer
  Vision, pp. 356--369 (2010)

\bibitem{Tombari2010Uniqueorignal}
Tombari, F., Salti, S., Stefano, L.D.: Unique signatures of histograms for
  local surface description.
\newblock Lecture Notes in Computer Science \textbf{6313}, 356--369 (2010)

\bibitem{Tombari2011A}
Tombari, F., Salti, S., Stefano, L.D.: A combined texture-shape descriptor for
  enhanced 3d feature matching \textbf{263}(4), 809--812 (2011)

\bibitem{Triebel2006}
Triebel, R., Kersting, K., Burgard, W.: Robust 3d scan point classification
  using associative markov networks.
\newblock In: Proceedings 2006 IEEE International Conference on Robotics and
  Automation, 2006. ICRA 2006., pp. 2603--2608 (2006).
\newblock \doi{10.1109/ROBOT.2006.1642094}

\bibitem{Vandapel2004}
Vandapel, N., Huber, D.F., Kapuria, A., Hebert, M.: Natural terrain
  classification using 3-d ladar data.
\newblock In: Robotics and Automation, 2004. Proceedings. ICRA '04. 2004 IEEE
  International Conference on, vol.~5, pp. 5117--5122 (2004).
\newblock \doi{10.1109/ROBOT.2004.1302529}

\bibitem{Wahl2003Surflet}
Wahl, E., Hillenbrand, U., Hirzinger, G.: Surflet-pair-relation histograms: A
  statistical 3d-shape representation for rapid classification.
\newblock In: International Conference on 3-D Digital Imaging and Modeling,
  2003. 3dim 2003. Proceedings, pp. 474--481 (2003)

\bibitem{wang2018local}
Wang, C., Samari, B., Siddiqi, K.: Local spectral graph convolution for point
  set feature learning.
\newblock In: Proceedings of the European Conference on Computer Vision (ECCV),
  pp. 52--66 (2018)

\bibitem{wang2015voting}
Wang, D.Z., Posner, I.: Voting for voting in online point cloud object
  detection.
\newblock In: Robotics: Science and Systems, vol.~1, pp. 10--15607 (2015)

\bibitem{wang2019graph}
Wang, L., Huang, Y., Hou, Y., Zhang, S., Shan, J.: Graph attention convolution
  for point cloud semantic segmentation.
\newblock In: Proceedings of the IEEE Conference on Computer Vision and Pattern
  Recognition, pp. 10296--10305 (2019)

\bibitem{wang2017cnn}
Wang, P.S., Liu, Y., Guo, Y.X., Sun, C.Y., Tong, X.: O-cnn: Octree-based
  convolutional neural networks for 3d shape analysis.
\newblock ACM Transactions on Graphics (TOG) \textbf{36}(4), 72 (2017)

\bibitem{wang2018deep}
Wang, S., Suo, S., Ma, W.C., Pokrovsky, A., Urtasun, R.: Deep parametric
  continuous convolutional neural networks.
\newblock In: Proceedings of the IEEE Conference on Computer Vision and Pattern
  Recognition, pp. 2589--2597 (2018)

\bibitem{wang2018sgpn}
Wang, W., Yu, R., Huang, Q., Neumann, U.: Sgpn: Similarity group proposal
  network for 3d point cloud instance segmentation.
\newblock In: Proceedings of the IEEE Conference on Computer Vision and Pattern
  Recognition, pp. 2569--2578 (2018)

\bibitem{wang2019associatively}
Wang, X., Liu, S., Shen, X., Shen, C., Jia, J.: Associatively segmenting
  instances and semantics in point clouds.
\newblock arXiv preprint arXiv:1902.09852  (2019)

\bibitem{wang2018dynamic}
Wang, Y., Sun, Y., Liu, Z., Sarma, S.E., Bronstein, M.M., Solomon, J.M.:
  Dynamic graph cnn for learning on point clouds.
\newblock arXiv preprint arXiv:1801.07829  (2018)

\bibitem{Wang2015A}
Wang, Z., Zhang, L., Fang, T., Mathiopoulos, P.T., Tong, X., Qu, H., Xiao, Z.,
  Li, F., Chen, D.: A multiscale and hierarchical feature extraction method for
  terrestrial laser scanning point cloud classification.
\newblock IEEE Transactions on Geoscience and Remote Sensing \textbf{53}(5),
  2409--2425 (2015)

\bibitem{Wohlkinger2012Ensemble}
Wohlkinger, W., Vincze, M.: Ensemble of shape functions for 3d object
  classification.
\newblock In: IEEE International Conference on Robotics and Biomimetics, pp.
  2987--2992 (2012)

\bibitem{wu2019pointconv}
Wu, W., Qi, Z., Fuxin, L.: Pointconv: Deep convolutional networks on 3d point
  clouds.
\newblock In: Proceedings of the IEEE Conference on Computer Vision and Pattern
  Recognition, pp. 9621--9630 (2019)

\bibitem{wu20153d}
Wu, Z., Song, S., Khosla, A., Yu, F., Zhang, L., Tang, X., Xiao, J.: 3d
  shapenets: A deep representation for volumetric shapes.
\newblock In: Proceedings of the IEEE conference on computer vision and pattern
  recognition, pp. 1912--1920 (2015)

\bibitem{xie2018attentional}
Xie, S., Liu, S., Chen, Z., Tu, Z.: Attentional shapecontextnet for point cloud
  recognition.
\newblock In: Proceedings of the IEEE Conference on Computer Vision and Pattern
  Recognition, pp. 4606--4615 (2018)

\bibitem{xu2020weakly}
Xu, X., Lee, G.H.: Weakly supervised semantic point cloud segmentation: Towards
  10x fewer labels.
\newblock In: Proceedings of the IEEE/CVF Conference on Computer Vision and
  Pattern Recognition, pp. 13706--13715 (2020)

\bibitem{xu2018spidercnn}
Xu, Y., Fan, T., Xu, M., Zeng, L., Qiao, Y.: Spidercnn: Deep learning on point
  sets with parameterized convolutional filters.
\newblock In: Proceedings of the European Conference on Computer Vision (ECCV),
  pp. 87--102 (2018)

\bibitem{Yang2016A}
Yang, J., Cao, Z., Zhang, Q.: A fast and robust local descriptor for 3d point
  cloud registration.
\newblock Information Sciences \textbf{s 346–347}, 163--179 (2016)

\bibitem{yang2018foldingnet}
Yang, Y., Feng, C., Shen, Y., Tian, D.: Foldingnet: Point cloud auto-encoder
  via deep grid deformation.
\newblock In: Proceedings of the IEEE Conference on Computer Vision and Pattern
  Recognition, pp. 206--215 (2018)

\bibitem{yi2016scalable}
Yi, L., Kim, V.G., Ceylan, D., Shen, I., Yan, M., Su, H., Lu, C., Huang, Q.,
  Sheffer, A., Guibas, L., et~al.: A scalable active framework for region
  annotation in 3d shape collections.
\newblock ACM Transactions on Graphics (TOG) \textbf{35}(6), 210 (2016)

\bibitem{Zelener2015Classification}
Zelener, A., Mordohai, P., Stamos, I.: Classification of vehicle parts in
  unstructured 3d point clouds.
\newblock In: International Conference on 3d Vision, pp. 147--154 (2015)

\bibitem{zeng20183dcontextnet}
Zeng, W., Gevers, T.: 3dcontextnet: Kd tree guided hierarchical learning of
  point clouds using local and global contextual cues.
\newblock In: European Conference on Computer Vision, pp. 314--330. Springer
  (2018)

\bibitem{zhang2020pointfilter}
Zhang, D., Lu, X., Qin, H., He, Y.: Pointfilter: Point cloud filtering via
  encoder-decoder modeling.
\newblock arXiv preprint arXiv:2002.05968  (2020)

\bibitem{Zhao2014Height}
Zhao, G., Yuan, J., Dang, K.: Height gradient histogram (high) for 3d scene
  labeling.
\newblock In: International Conference on 3d Vision, pp. 569--576 (2014)

\bibitem{zhao2019pointweb}
Zhao, H., Jiang, L., Fu, C.W., Jia, J.: Pointweb: Enhancing local neighborhood
  features for point cloud processing.
\newblock In: Proceedings of the IEEE Conference on Computer Vision and Pattern
  Recognition, pp. 5565--5573 (2019)

\bibitem{Zhong2010Intrinsic}
Zhong, Y.: Intrinsic shape signatures: A shape descriptor for 3d object
  recognition.
\newblock In: IEEE International Conference on Computer Vision Workshops, pp.
  689--696 (2010)

\bibitem{zhou2018learning}
Zhou, L., Zhu, S., Luo, Z., Shen, T., Zhang, R., Zhen, M., Fang, T., Quan, L.:
  Learning and matching multi-view descriptors for registration of point
  clouds.
\newblock In: Proceedings of the European Conference on Computer Vision (ECCV),
  pp. 505--522 (2018)

\bibitem{zhou2018voxelnet}
Zhou, Y., Tuzel, O.: Voxelnet: End-to-end learning for point cloud based 3d
  object detection.
\newblock In: Proceedings of the IEEE Conference on Computer Vision and Pattern
  Recognition, pp. 4490--4499 (2018)

\end{thebibliography}

\end{document}